*Draft of the paper appearing in the Italian Journal of Linguistics*
*37.1 (2025), p. 3-44 DOI: 10.26346/1120-2726-237*

# Is it the end of (generative) linguistics as we know it?

Cristiano Chesi

NeTS-IUSS Pavia

*Abstract*

A significant debate has emerged in response to a paper written by Steven Piantadosi (Piantadosi 2023) and uploaded to the LingBuzz platform, the open archive for generative linguistics. Piantadosi's dismissal of Chomsky's approach is ruthless, but generative linguists deserve it. In this paper, I will adopt three idealized perspectives—computational, theoretical, and experimental—to focus on two fundamental issues that lend partial support to Piantadosi's critique: (a) the evidence challenging the Poverty of Stimulus (PoS) hypothesis and (b) the notion of simplicity as conceived within mainstream Minimalism. In conclusion, I argue that, to reclaim a central role in language studies, generative linguistics—especially Minimalism, which represents a prototypical theoretical perspective on language—needs a serious update leading to (i) more precise, consistent, and complete formalizations of foundational intuitions and (ii) the establishment and utilization of a standardized dataset of crucial empirical evidence to evaluate the theory's adequacy. On the other hand, ignoring the formal perspective leads to major drawbacks in both computational and experimental approaches. Neither descriptive nor explanatory adequacy can be easily achieved without the precise formulation of general principles that can be challenged empirically.

## *1. Introduction*

In November 2013, a group of prominent formal linguists[1] was invited by Andrea Moro to the University School for Advanced Studies IUSS Pavia to discuss foundational issues of generative grammar. The bombastic title of the roundtable I helped organize was "Can there be a Hilbert List of Syntax (yet)?". A few months before, Noam Chomsky delivered an inspiring opening lecture for the 2012 academic year at IUSS on "Language and the Limits of Understanding". Stimulated by his insightful commentary, we approached the roundtable with considerable excitement, ready to delve into the current state of generative linguistics and to outline its history of success. Just to mention three highly influential ideas whose proposers were present at that roundtable, we could list: (a) The featural sensitivity of locality, known as *Relativized Minimality* (Rizzi 1990); a very pervasive and persuasive restriction on specific kinds of non-local dependencies with a clear impact not only on the elegance of the grammatical competence description, but also on certain performance issues—e.g., under the insightful interpretation of relativized minimality in terms of featural inclusion (Starke 2001), the difficulty in processing object relative clauses can be attributed to the challenge of differentiating distinctive features (Friedmann, Belletti, and Rizzi 2009; Grillo 2008). (b) The universal ordering of functional categories (Cinque 1999; Rizzi 1997), despite their apparent cross-linguistic variation; a study that originated a fruitful inquiry program dubbed *Cartography* (Belletti, 2004; Cinque, 2002; Rizzi, 2004; Rizzi & Cinque, 2016 a.o.). (c) The complicated—but stable—relationship between hierarchy and linear order, as defined by *Antisymmetry* (Kayne 1994); a formal intuition that restricts the number of plausible structural descriptions logically admitted on the basis of a dominance-precedence mapping concern. If the number of options for pairing hierarchical structure with the linearly ordered set of pronounced (or signed) morphemes—necessarily imposed by the sensory-motor system—were restricted, the "logical problem of language acquisition" would be more manageable and potentially more solvable under the *Principles and Parameters* perspective (Noam Chomsky 1981).

[1] Adriana Belletti, Guglielmo Cinque, Denis Delfitto, Marcel Den Dikken, Anna Maria Di Sciullo, Robert Frank, Hubert Haider, Richard Kayne, Giuseppe Longobardi, Rita Manzini, Norvin Richards, Henk van Riemsdijk, Luigi Rizzi, Ian Roberts, Dominique Sportiche, and Peter Svenonius.

At the end of two days of discussion, after 16 brilliant talks, we tried to take stock of the proposed problems agenda, but our attempt was rather disappointing: although each problem was apparently crucial and well stated, the extension of the relevant empirical basis fitting the specific theory set-up was sometimes hard or faint. More crucially, it was practically impossible to present all the problems in a concise and coherent manner within a consistent framework: in nearly every instance, although each problem statement came with a proposed solution, the underlying assumptions were often at odds with the premises of others.

A spectre was haunting generative linguistics—the spectre of Minimalism. In fact, this turned out to be a sum of idiosyncratic interpretations.

In Chomsky's intuition, Occam's razor was needed to purge the evolution of *Government and Binding Theory* (Noam Chomsky 1981) of unnecessary machinery. As a result, the theory should have been simpler, mathematically sound, computationally efficient, and include nothing but what is strictly necessary. This approach marked the beginning of a new research program dubbed Minimalism (Noam Chomsky 1995). Thirty years on, it must be acknowledged that while the program began with commendable intentions, the emerging framework still lacks consistency, especially regarding the application of fundamental structure-building operations to empirical problems.

In the end, events have surpassed the intentions of the roundtable. A thorough investigation into the success of very Large Language Models (vLLMs)[2], alongside statistical and experimental advancements, may lead one to agree with many in concluding that generative linguistics no longer dictates the agenda for future linguistic challenges. This position is summarized by Steven Piantadosi who fosters the idea that vLLMs are "genuine theories of language, including representations of syntactic and semantic structure" (Piantadosi 2023). Although these models are primarily designed for a wide range of Natural Language Processing tasks, from Machine Translation to Question Answering, Piantadosi's key argument is their superiority as linguistic theories. According to him, these models surpass generative approaches in performing comprehensive syntactic tests, ranking vLLMs as the most effective linguistic theories currently available. One platform designed for performing such linguistic benchmarks is SyntaxGym (Hu et al. 2020): an on-line, open-source repository that includes a significant set of linguistic contrasts—39 test suites that include a total of about 4k sentences. For each relevant contrast included, human generalizations have been gathered in various studies. Direct comparisons of these data with the predictions provided by the models under evaluation is then possible. An exemplificative contrast included in SyntaxGym involves non-local agreement dependency, where the subject must agree in number with the matrix copula, despite being linearly separated by a relative clause (from Marvin & Linzen, 2018):

(1) a. *The author* that the senators hurt *is* good
b. **The author* that the senators hurt *are* good

[2] The full list of abbreviations used is provided at the end of the paper.

A model predicting that (1).a "is better than" (1).b represents a more adequate theory with respect to a theory that is not able to infer the ungrammaticality of (1).b. The utility of these minimal pairs in linguistic theorizing is uncontroversial and can be further sophisticated. In the example below, for instance, another kind of restriction on non-local dependencies is considered, also known as Across-The-Board extraction, ATB (Williams 1977):

(2) a. I know *what*$_i$ the guy broke _$_i$ accidentally and the mechanic fixed _$_i$ skilfully.
b. *I know *what*$_i$ the guy broke _$_i$ accidentally and the mechanic fixed *the engine* skilfully.

This ATB constraint predicts that when a wh- item ("what") is extracted from the first conjunct ("the guy broke _ accidentally"), a gap coindexed with the same wh- item should be present also in the second conjunct ("the mechanic fixed _ skilfully"). The relevant contrast, again included in SyntaxGym, compares the correct configuration (2).a with an ungrammatical minimal variation in which the second gap is filled by an intrusive argument ("the engine"), in (2).b.

Although the Minimalism framework is explicitly designed to address these issues, paradoxically, I concur with Piantadosi's critique regarding its inability (at least in mainstream set-ups) to perform adequately in similarly complete and extensive benchmarks as the ones presented in SyntaxGym[3].

To frame this problem, we need to consider three major perspectives, albeit somewhat idealized for the sake of discussion: first, the computational perspective, which posits that the best linguistic theory is simply the one that performs optimally on a shared test set. Second, the theoretical perspective, that considers this goal (i.e., observational adequacy) just as a starting point, being the final goals descriptive adequacy—the theory should be grounded in robust "genuine generalizations" (Noam Chomsky 2021)—and explanatory adequacy—the theory should account for language learnability. Lastly, the experimental perspective, that underscores the imperative of meticulous data collection and analysis, signaling a departure from purely theoretical "armchair linguistics". I will contend that, unless integrated, these perspectives individually lead to impasses.

In fact, all these three perspectives will be essential to reexplore two foundational issues (section §3): the Poverty of Stimulus (PoS) hypothesis (§3.1) and the notion of simplicity applied to structure-building operations (Merge and Move) in mainstream Minimalism (§3.2). In both cases, I will provide logical and empirical evidence suggesting that the classic arguments must be re-worked, and the three perspectives are all necessary to avoid confounds.

[3] In fact, a growing body of research suggests that the performance of state-of-the-art vLLMs falls short of human-level morphosyntactic competence. We will revisit this critical issue in §3.1 (Chesi, Vespignani, and Zamparelli to appear; Dentella, Günther, and Leivada 2023).

My primary concern is that the Minimalist Program's underspecification of key concepts, including simple but effective restrictions on the application of structure-building operations, has become untenable (formalization issue). Furthermore, the general underevaluation of experimental and computational advancements by leading scholars in the generative field has contributed to a perception of generative linguistics as marginal within both computational and experimental language research communities. I will argue here that it is imperative to bridge the formalization gaps as effectively as possible and to adopt a modern approach to theory evaluation that relies on shared datasets and metrics (evaluation issue). On the other hand, what effectively guides sound (linguistic) inquiry is the search for empirical evidence that rejects a specific theoretical setup, while avoiding confirmation bias as much as possible. In this respect, generative linguists benefit from both valuable experience and clear principles which, once explicitly formalized, can be disproved. In my opinion, this remains the only effective way to build descriptively adequate theories.

To address these issues, the next section (§2) will set the stage by defining the core minimalist concepts and the empirical data that modern methods have made available. I will first introduce the "Language Problem" that any theory must confront. Subsequently, I will discuss the widely adopted "T-model" (Noam Chomsky 1981; Noam Chomsky et al. 2023) and examine the criteria for achieving descriptive adequacy and computational efficiency. I will then explore a four-way classification of relevant empirical data (§2.2.1-§2.2.4). This will pave the way towards a re-analysis of the PoS argument, emphasizing that the only relevant source of linguistic information for explaining language acquisition, and thereby targeting explanatory adequacy, remains positive linguistic evidence found in child-directed speech. While assessing a theory's descriptive adequacy requires all possible experimental evidence, from the learnability perspective, any implicit metalinguistic information is deemed irrelevant.

## 2. *Empirical Evidence for a Theoretical Perspective*

### 2.1. *Setting the Stage: From Descriptive Adequacy to Efficiency Considerations*

The primary goal of any linguistic theory (X) is to precisely circumscribe the infinite set (language L) composed by those sentences (Ss) judged as grammatical by native speakers. By *sentence* we simply refer to a (compositionally) interpretable and producible (through signs or sounds) ordered string[4] of words/morphemes. On this basis, we can define the Language Problem:

**Definition 1.** *Language Problem*
Is theory X capable of generating and recognizing all and only the sentences Ss belonging to language L?

[4] It is usually (anecdotally) assumed that sign languages pose a less strict requirement on lexical items linearization. The actual amount of simultaneity empirically observed in specific constructions, across sign languages is, in fact, precisely documented (Vermeerbergen, Leeson, and Crasborn 2007).

A *Minimalist framework* is a theoretical perspective promoting the shortest possible list of *instructions*[5] that would enable a grammatical theory to solve the Language Problem, for any natural language L. A theory that solves the Language Problem is considered *observationally adequate* (Noam Chomsky 1964).

In structural terms—that is, an abstract and explicit description of the generalizations based on the observed Ss in L—, a *Minimalist Grammar* (*MG*) defines an infinite set of derivations (sequences of steps, $D_S$) obtained through the applications of essentially one simple structure building operation (*Merge*) over lexical items ($l_i$) selected from the language lexicon ($Lex_L$), as exemplified in (3):

(3) Derivation ($D_s$) of the sentence S: "Alice scolds Bill"
i. *Select*(Alice, Bill, scolds) where {Alice, Bill, scolds} $\in Lex_{English}$
ii. *Merge*(scolds, Bill) = {scolds, Bill}
iii. *Merge*({scolds, Bill}, Alice) = {Alice, {scolds, Bill}}

We usually adopt a concise syntactic tree-like structural description, $T_S$ in (4), to represent the history of the derivation $D_S$ presented in (3).

(4) Syntactic representation ($T_S$) of S as derived in (3)

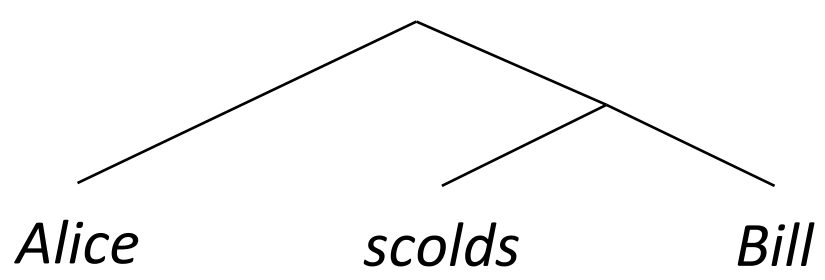

Following the generative grammar tradition, a MG is a theory of I(nternal)-language, modeling the competence of an idealized speaker (Noam Chomsky 1965, 11:24): this minimally comprises a lexicon specification ($Lex_L$) and the procedure to obtain appropriate derivations resulting in structural descriptions. Those descriptions should be coherent with the way we pronounce—e.g., /ˈælɪs skəʊldz bɪl/—and understand—e.g., "Alice" is the agent X, "Bill" is the patient Y, "scolds" the predicate, then "X scolds Y"—the corresponding sentences Ss in L—e.g., "Alice scolds Bill". In this sense, $T_S$ should be a legible scaffolding both for a *Phonetic Form*, *PF*, at the *Sensory-Motor* interface *SM*, and for interpretation in terms of *Logical Form*, *LF*, a compositionally legible structure at the *Conceptual-Intentional* interface, *CI*.

Once a pronounceable/interpretable fragment of phrase structure is created—using the structure building operation Merge in a workspace, *WP*—, this fragment is delivered—"spelled-out"—to the two interfaces: CI on the one side and SM on the other. The core syntactic engine will continue to build structures, while the two external modules will independently elaborate on those representations.

These components can be organized under the standard "T-" (or "Y-") model elaborated in (5). Notice that the observable Ss sentences—E(xternal)-language—forming the empirical restriction of the L set are placed at PF, which is the sole stage at which we can empirically observe them. Crucially, the idealized L identified as I-Language, plus externalization constraints (e.g. linear order at PF), should exactly correspond to that observable E-language L.

---

[5] I have deliberately used the term "instructions" to include both the classic notion of (parametrized) principles—such as *X'-theory*, *theta-criterion* or *case filter* (Noam Chomsky 1981)—and structure building operations—such as *Merge* or *Move*, (Noam Chomsky 1995).

(5) The "T-model" (re-adapted and extended from Chomsky, Seely, et al., 2023)

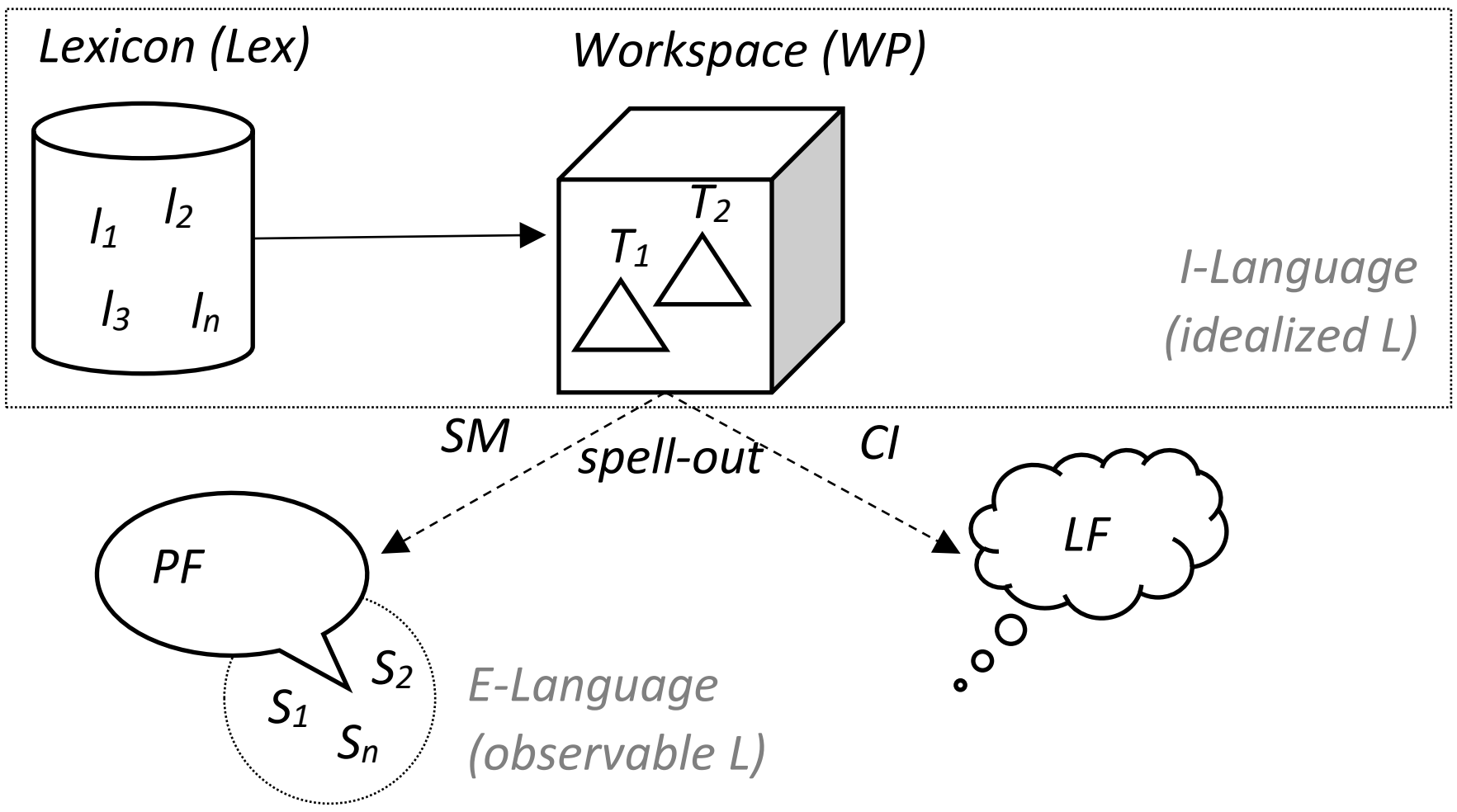

The Minimalist Program was the result of a slimming diet for grammatical theories that appeared to be necessary, at the beginning of the '90s, due to two independent reasons: the first reason was related to the idea of "perfection" (Noam Chomsky 1995, 1), the second to the "proliferation of unwanted derivations" (Noam Chomsky 1995, 283).

As far as the notion of "perfection" is concerned, looking for a theory that only includes the strictly necessary assumptions to satisfy the interface requirements (at SM and CI) is simply restating a long-standing logical problem formulated in many equivalent ways. For instance, in Minimum Description Length, MDL, terms (Grünwald 2007; Rissanen 1978; 1987), or Solomonoff-Kolmogorov-Chaitin complexity (Chaitin, 1969; Kolmogorov, 1963; Solomonoff, 1960), the best theory for a specific empirical domain is the one that better compresses the data observed in that domain. Despite an undecidability menace (M. Li and Vitányi 2008), this is a practical mathematical way to compare the *descriptive adequacy* fit (Noam Chomsky 1965) of a theory (a step forward with respect to *observational adequacy*) that subsumes the notion of *simplicity* (Noam Chomsky 2021):

**Definition 2.** *Descriptive adequacy*
the theory that captures more data (Ss in L) with fewer instructions (driving $D_S$ to derive $T_S$) is the one descriptively more adequate.

The second reason was related to computability: a linguistic theory must be computationally efficient, that is, we should prefer (computable) theories that perform a relevant derivation in the most economical way (Noam Chomsky 1995, 8–9)[6].

[6] One might be tempted to add "without generating unwanted structures (Frampton and Gutmann 2002)" but this is already implicit in Definition 3: a theory Y that performs a wrong derivation D2, in addition to the appropriate one D1, is less efficient with respect to the theory X that only performs D1.

**Definition 3.** *Computational efficiency*
the computationally most efficient derivation D is the one obtaining the intended structure T in fewer steps.

The distinction between "fewer steps" and "fewer instructions" is critically important: the former refers to the *time complexity* of an algorithm (Barton, Berwick, and Ristad 1987), while the latter pertains to the dimensions of the theory. To understand time complexity, assume a serial procedure where each operation is performed sequentially. Considering a hypothetical scenario in which a very slow processor is capable of executing just one operation per second, and assuming that both Select and Merge count as single operations, executing the procedure (3) would require exactly three seconds. "Fewer instructions", on the other hand, refers to the size of the theory which, in simple terms, can be expressed by the number of bits required to encode the lexicon, the structure-building operations, and the procedure driving the derivations. For example, in Python, less than 300 bytes are needed to encode the instructions for the derivation described in (3)[7].

A generative linguist might feel uncomfortable with these definitions and try to back off under the *competence-performance* shield: the goal of any linguistic theory should be restricted to defining the domain of computation, known as the competence level—or computational level, in Marr's terminology (Marr 1982)—, and should not encompass the performance level—typically associated with Marr's algorithmic level. This defense falls short in two ways: first, even though the sequence of operations applied to obtain a certain derivation is assumed to be abstract, with no real temporal implications whatsoever (Noam Chomsky 1995, 380), computational efficiency is included in any MG version—e.g. the *phase* idea[8] (Noam Chomsky 2001); this implies that algorithmic complexity is at issue here with no necessity to invoke performance. Second, considering the classic representational vs. derivational opposition[9] one might still prefer *representational filters* on a complete phrase structure T, once it is fully built—e.g., C-command constraints—, but this approach is logically disadvantaged compared to any derivational constraint that would exclude the possibility of deriving unwanted structures Ds well before those ill formed descriptions will be later, representationally, valued then discarded to obtain T. This seems to me a straightforward argument on the superiority of derivational approaches over representational ones in computational terms (cf. footnote 6). One might be suspicious about the mere numeric comparison suggested by Definition 2, but we will see soon that this comparison can be safely restricted to truly recursive generalizations all equally able to generate and recognize an infinite set of data—e.g., relative clause formation.

[7] This is the size needed for a lexicon of just three words. Expanding the lexicon to 300 words, the code size will increase to approximately 5.7 KB. For a more extensive lexicon of 300,000 tokens, about 5.5 MB will be necessary. It is important to note that these lexical entries do not include any featural specifications and Select and Merge operations are completely unconstrained. For comparison, a vLLM such as GPT-3 requires about 350GB once optimized (Radford et al. 2018).

[8] A "phase" is a derivational chunk intended to minimize the computational burden. During the derivation, the items selected from the lexicon are arranged and sent to the interfaces—spelled-out—phase-by-phase. An uncontroversial phase—spell-out—domain is the Complementizer Phrase (CP), that is, embedded sentences are assembled and spelled-out before the matrix one.

[9] A representational perspective only defines the relevant constraints in terms of the geometry of the syntactic tree. A classic representational constraint that applies to gaps/traces licensing is *C(onstituent)-command*: "Node A C-(onstituent)-commands node B if neither A nor B dominates the other and the first branching node which dominates A dominates B" (Reinhart 1976, 8). From a representational perspective the order in which single phrases are assembled is irrelevant. Conversely, this order is crucial in derivational approaches—MGs: from this perspective the order in which structure-building operations apply is fundamental to derive phrase structures. As a consequence, filters such as C-command must be reformulated in derivational terms, e.g., "X C-commands all and only the terms of the category Y with which X was paired/concatenated by Merge or by Move in the course of the derivation" (Epstein et al. 1998, 32).

To summarize, descriptive adequacy (Definition 2) and computational efficiency (Definition 3) pose theoretical boundaries to the Language Problem (Definition 1). While the theoretical perspective tends to idealize the empirical domain to enhance descriptive adequacy (I-language), this approach faces two challenges: firstly, experimental methods tend to enrich the empirical domain in ways that do not always align with idealized contrasts; secondly, the computational perspective, enhanced by machine learning methods, is naturally suited to narrow the gap between idealized language and the observed language L, efficiently and effectively processing any observed sentence S. Let us explore these challenges in more detail in the next section.

### *2.2. Competence, Performance, and Explanatory Adequacy*

After the cognitive revolution (Noam Chomsky 1959), a precise method emerged to approach the study of high cognitive functions. This method was used to avoid imprecise theoretical notions and to obtain fruitful generalizations from the limited set of observable/available data. Under the definitions we just provided in §2.1, one might well imagine that our language faculty be not so perfect—as correctly programmatically assumed in principle by Minimalism. In this case, the computational search for perfection might lead us to solutions to the Language Problem that are more performant than the ones nature has provided.

A classic example is motion: we quickly realized that wheels were more efficient than legs for moving an object from position A to position B. “More efficient” can be interpreted in terms of faster transitions from A to B, more robust—in terms of failures during the motion—, or less complex solutions—simpler design and control. A “minimalist program” for human motion, focusing solely on computational efficiency, would have favored wheels over legs[10]. For this reason, the Language Problem (Definition 1) must be coupled with another fundamental constraint to limit the greediness of Computational Efficiency (Definition 3): we do not want to derive random structure—mutatis mutandis: any possible motion planning from A to B in arbitrary conditions—, but just those structures that are relevant descriptions for the sentences in our language L. From the perspective of the “motion problem”, the solution of using wheels would not be suitable for tasks like climbing stairs, which humans, in fact, perform quite well under standard circumstances. At the same time, the temporal predictions of transitions from A to B will be greatly underestimated—wheel-based vehicles would take much less time to move from A to B on a highway than humans.

Even though computational efficiency is, per se, a legitimate independent goal, as cognitive scientists we are primarily interested in understanding our language faculty—I-language description—, that is, producing a theory that is, at least, descriptively adequate (Definition 2). Therefore, a trade-off between descriptive adequacy and computational efficiency emerges, possibly favoring a theory that is less compact or efficient but fits better with the data associated with the relevant derivations of sentences Ss in L.

[10] This is in fact a pervasive solution to the motion problem in robotics (Nosengo 2014).

We usually refer to *Performance* as the actual usage of our linguistic knowledge (Noam Chomsky 1965), that is, the way native speakers perform a recognition/generation of sentences Ss in their language L. Ideally, this performance should be precisely predicted by applying externalization considerations—e.g., linear order on structured lexical items—and specific resources constraints—e.g., memory or time limitations—to the derivations obtained from the descriptively adequate linguistic theory X, which is nothing but a formal description of our linguistic *Competence*. The distinction between Competence and Performance is traditionally viewed as logical and well-defined[11]. However, it is important to note that performance is the only empirical data source available to us for defining the boundaries of our language L. Thus, it is useful to introduce a four-way classification of performance data, independently considering the (i) meta-linguistic, (ii) categorical, (iii) temporal, and (iv) naturalistic dimensions. These dimensions, carved into the experimental perspective tradition, will enable us to frame the Language Problem (Definition 1) more effectively.

### *2.2.1. Meta-linguistic Dimension*

Linguistic judgments can be *explicit*—meta-linguistic—or *implicit*: we might ask a native speaker to judge, overtly, a sentence for grammaticality—or complexity, or acceptability—or ask the same speaker to simply read or repeat a sentence. In the first case, the speaker must access its meta-linguistic knowledge and provide a judgment that is otherwise unconscious unless overtly probed. In the second case, we can consider implicit measures like reading-times or errors in the very same contexts.

### *2.2.2. Categorical Dimension*

An observed linguistic behavior can be *categorical*—binary—or *gradual*: if we ask for grammaticality judgments, a categorical judgment would be a binary choice between grammatical or ungrammatical, with no third option. The same judgment might be asked on a Likert scale from 1 (fully ungrammatical) to 7 (perfectly grammatical). An elicited answer to a specific linguistic question might also be considered categorical—correct or wrong—or gradual—receiving a score from 1 to 5, or similar. Although for a reasonable amount of repeated observations and a sufficient number of participants, these two measures converge (Sprouse and Almeida 2017), it remains a deep theoretical question to accept gradual judgments as a source of relevant empirical evidence for a theory. Considering, for instance, the generative power of a grammar (Chomsky, 1956), categorical judgments are the only relevant empirical data we can consider.

### *2.2.3. Temporal Dimension*

We can observe/elicit a linguistic behavior at the end of sentence processing (*off-line*) or during sentence processing (*on-line*): in the first case, we might ask a comprehension question or a grammaticality judgment when the exposure to the linguistic input to judge is completed. In the second case, we can record implicit—reading times in self-paced reading, fixation times in eye-tracking, neurometabolic responses recorded using EEG or fMRI, etc.—or explicit—e.g., maze techniques, Forster et al., 2009—measures while the linguistic input under evaluation is processed, morpheme by morpheme, word by word or region by region.

[11] A closer look might reveal logical inconsistencies: see the Competence Paradox, (Chesi and Moro 2015).

### *2.2.4. Naturalistic Dimension*

Naturalistic data are those collected using natural linguistic stimuli—e.g., narrative texts or public speeches—in *ecological* conditions—e.g., reading or listening. This data collection approach, grounded in a solid corpus linguistic tradition, is complementary to the *experimental* one, in which the linguistic input is accurately manipulated using minimal pairs and the task fully controlled to selectively evaluate the contribution of a specific variable in linguistic processing. Naturalistic/ecological studies are becoming very popular and are now widely considered useful sources of implicit processing information (Brennan et al. 2016; Siegelman et al. 2022). On the other hand, experimental practice on controlled inputs has significantly advanced our understanding of language competence by targeting the influence of specific variables on minimal contrasts.

### *2.2.5. Between Experimental and Computational Linguistics*

Employing a precise typology of empirical data assists us in reframing the tension between descriptive adequacy and computational efficiency, as well as clarifying the division of labor between competence and performance.

A standard practice, dating back to at least Chomsky and Miller's explanation of why native (naïve) speakers accept certain linguistic forms, (6).a, but reject other structurally similar ones, (6).b, is fundamentally tied to the competence/performance divide (Miller and Chomsky 1963):

(6) a. (I saw) [a dog [that bit [a cat [that chased [a mouse that ran away]]]]].
b. (I saw) [a mouse [that [a cat [that [a dog bit]] chased]] ran away].

A theory X, which predicts that restrictive relative clauses can modify indefinite Determiner Phrases (DPs), faces difficulty in restricting this (recursive) operation to non-nested contexts, such as (6).a: opting for a solution that relies on an independent performance (algorithmic) domain, where available resources are limited and exhausted in (6).b but not in (6).a, appears to be an appealing approach. As a result, our competence/computational perspective will be essentially declarative and make predictions on empirical data which are only (i) explicit (grammaticality judgments), (ii) categorical (grammatical vs. ungrammatical), and (iii) off-line. Moreover, such an approach can hardly take advantage of naturalistic/ecological data while single grammatical rules/principles can be more easily tested under (iv) controlled conditions. While this option is tenable and logical, we can demonstrate that this solution is less descriptively adequate (Definition 2) when compared to a theory that can derive ($D_S$) a relevant structure ($T_S$) for a given sentence (S) with fewer assumptions, simply by operating at the algorithmic level.

Notice that, if we do not fully rely on native speaker judgments and we “idealize” them, we widen the gap between the language as it's truly observed and the idealized one. By doing so, we implicitly adopt the following principle:

**Definition 4.** *The dust under the carpet principle*
If a theory X *overgeneralizes* or *undergeneralizes* on sentence S, we can maintain X by excluding the data related to performance on S.

i. a theory *overgeneralizes* when it predicts a relevant structure $T_S$ for S through derivation $D_S$, but native speakers judge S as ungrammatical because they cannot perform $D_S$;

ii. a theory *undergeneralizes* when it is unable to predict a structure $T_S$ for S through derivation $D_S$, but native speakers judge S as grammatical by performing $D_S$.

Everything being equal, it must be recognized that our descriptive adequacy definition suggests that a theory Y that can "remove the dust from under the carpet" will be descriptively more adequate than X. Similarly, a theory Z that expands the empirical domain, taking advantage of many data sources, everything being equal, will be superior to X: by Definition 2, the theory that is able to generalize—i.e., capturing more data with fewer instructions—on more (i) explicit and implicit, (ii) categorical and gradual, (iii) off-line and online (iv) ecological and controlled data sources is descriptively more adequate.

In the end, the adopted notion of descriptive adequacy has the considerable advantage of being measurable: obviously, as clearly stated by Chomsky (Noam Chomsky 2021), we can not simply rely on numerical comparisons since also the most trivial recursive grammar will be able to predict an infinite set of sentences/data. On the other hand, there is no need to weaken the idea that an "appropriate generalization" is simply "a shorter set of instructions" (or rules or principles) that capture an infinite set of (equivalent) constructions[12]. A theory that is equally descriptively adequate in capturing the same set of sentences at a comparable cost—number of "instructions", equivalent structural T descriptions—will be challenged by different measures we can possibly obtain on the very same sentences: if a theory Z is not only able to predict that both sentences $S_1$ and $S_2$ are grammatical but also to which extent $S_1$ is processed more easily than $S_2$, then theory Z must be considered descriptively superior compared to theory X which is not able to rank $S_1$ and $S_2$.

So far, we have observed that: (i) although linguistic data available can be richly characterized, not all theoretical approaches can effectively utilize them; (ii) the trade-off between computational efficiency (fewer steps) and descriptive adequacy (fewer instructions) may favor larger theories that capture a broader array of data types. Consequently, the theories that are more descriptively adequate tend to be those that are observationally more adequate—that is, those that simply predict a larger quantity of data. This plainly supports Piantadosi's point, which highlights the—observational—adequacy of vLLMs. However, in the end, this would lead to the inclusion of a vast number of irrelevant exceptions merely to broaden the empirical coverage of the theory. This approach would significantly disadvantage theories that cleverly apply "the dust under the carpet" principle to dismiss a long tail of idiosyncratic phenomena (e.g., those based on world knowledge rather than on linguistic competence).

[12] One might be tempted to provide a more explicit definition of "construction", as Chomsky has done by referring to "legitimate generalizations" (Noam Chomsky 2021). This is unnecessary: if a generalization $T_S$, obtained through $D_S$, is adequate—*legitimate*, in Chomsky's terms —, regardless of how it is formulated, it will capture an infinite set of irrelevant lexical variations of the sentence S; if it is inadequate—or *illegitimate* —, it will fail to capture any variations. The issue then becomes one of size: the smaller, the better. This is what descriptive adequacy fundamentally entails (Definition 2).

Before precisely addressing an effective method for identifying "idiosyncratic data" (§3.1.2), it is crucial to mitigate a potential overemphasis on diverse empirical sources. To achieve this, it becomes essential to introduce an additional layer of adequacy based on the concept of *learnability* (Noam Chomsky 1965):

**Definition 5.** *Explanatory adequacy*
A theory that uses only primary linguistic data available to children to select a descriptively adequate model for L is explanatorily adequate.

Under Definition 5, if a theory Z only relies on positive primary linguistic data—a list of naturalistic Ss—to identify the language L (that also includes Ss), while a theory X needs "unreasonable" data (as we will see in §3.1) to make the same predictions of Z, then Z is explanatorily adequate while X is not.

We are now ready to address two critical points with which generative grammar, as it is currently widely conceived in mainstream Minimalism, fails to comply. Those are related to, first, the formalism adopted and, second, a shared test set to be used as a benchmark.

*Formalization Issue*
If a theory is not fully explicit—i.e., formalized—there is no way to make precise predictions. For computational linguists accustomed to running their models on a computer, it is a well-known fact that no external oracle can ever fix a bug or a gap in the theory.

To the best of my knowledge, Edward Stabler was the first scholar to make a serious attempt at formally and successfully articulating a MG in a testable manner (Stabler 1997). His attempt was inspired by Chomsky's 1995 formulation, but he filled various gaps to make the model sound and complete (in a pre-theoretical sense). One such example of "filling the gap" was related to the definition of successful Merge and its impact on word order. Considering Merge as the fundamental operation that takes "a pair of syntactic objects ($SO_i$, $SO_j$) and replaces them by a new combined syntactic object $SO_{ij}$" (Noam Chomsky 1995, 226), considering its asymmetric nature (Chomsky, 1995, p. 246, either $SO_i$ or $SO_j$ will project) and the inclusiveness condition (nothing but what is in the lexicon appears on phrase structure, Chomsky, 1995, p. 249), Stabler formulated a feature-driven operation that strongly restricts the "free" Merge operation: since we must limit the exuberance of this operation, before relying on later filters that would reduce both descriptive adequacy and computational efficiency (under Definition 2 and Definition 3 respectively), feature checking is a suitable option: $\alpha$ and $\beta$, two independent syntactic objects, will (re-)merge successfully if and only if $\alpha$ has a *probe*/*select*/*interpretable*/*licensor* feature and $\beta$ has a *goal*/*base*/*uninterpretable*/ *licensee* feature[13]. In more explicit terms, considering *=x* a categorial probe associated with the head $\alpha$ and *x* a categorial goal associated with the complement $\beta$, then:

(7) $\text{M}\small{\text{ERGE}}(\alpha_{=x}, {}_{x}\beta) = \{\alpha\ \{\alpha_{\cancel{=x}}, {}_{\cancel{x}}\beta\}\}$

[13] I tried to include here all the relevant names which have been attributed to these features since (Noam Chomsky 1995). Stabler, in its original paper, used the *select* vs *base* terms for Merge, and *licensors* and *licensee* for Move. Here, I simplified a bit his formalism, for instance ignoring the directionality of the selection (to the right, =X vs. to the left, X=) and the strong/weak distinction (=X and =x, respectively).

In prose: the MERGE operation must take two items, $\alpha$ and $\beta$, that are characterized as the head ($\alpha$, since it bears the probe feature "*=x*", e.g., "scolds$_{=D}$") and the complement ($\beta$, since it has the probed categorial feature "*x*", e.g., "$_{D}$Bill"). The result of this—destructive, since features are deleted once the operation is successful—MERGE operation is a T representation expressed in set-theoretical terms—$\alpha$ dominates/includes the unordered set {$\alpha$, $\beta$}, that is, {scolds {scolds$_{\cancel{=D}}$, $_{\cancel{D}}$Bill}}. To obtain a predictable linear order, Stabler suggests that MERGE probes the first goal "to the right", while extra probes in $\alpha$ will be selected "to the left"—this creates argumental shells (Stabler 1997, 7), e.g., in (3), (4): {scolds {$_{\cancel{D}}$Alice, scolds$_{\cancel{=D}}$ {scolds$_{\cancel{=D}\ =D}$, $_{\cancel{D}}$Bill}}}; using angled brackets to indicate the linearization of words, the predicted outcome is: <Alice, scolds, Bill>. The formalism developed by Stabler sparked a dynamic yet niche debate, primarily confined to the realm of mathematical logic and largely overlooked by both computational and generative linguists. This oversight persisted despite a commendable effort to update the model, incorporating newer concepts like Agree and the *Workspace* idea, through a collaboration with Chris Collins (Collins and Stabler 2016). Unfortunately, this update received limited attention, with only a select group of scholars recognizing the importance of a fully formalized theory (Noam Chomsky et al. 2023).

Observe that these complex formalization practices are not mere mathematical curiosities. Instead, they represent the only tangible method for testing and refining linguistic theories. The original formulation of Merge lacked a crucial component, making the Language Problem (Definition 1) logically intractable. Without Stabler's proposal, this fundamental problem would have remained unresolved. Chomsky might regard this as a hybrid formalism that incorporates an externalization constraint—linear order. However, the inclusion of this constraint is a valid solution for generating sentences that would otherwise remain unobservable. Upon evaluation for descriptive adequacy, this methodology facilitates direct comparison with other formalisms, enabling their ranking based solely on descriptive adequacy.

Furthermore, it is only through the formalizations provided by Stabler and his colleagues that we have been able to ascertain the expressive power of MG (Michaelis 2001) and explore different strategies for parsing and incremental analysis (Stabler 2013). Unfortunately, full theory formalization remains a rare practice in Minimalism—with laudable exceptions (Ginsburg and Fong 2019).

*Evaluation Issue*
Another essential component for fruitfully addressing the Language Problem is the creation of a shared test/reference set that encompasses all the relevant contrasts we aim to capture. Again, this is standard practice in computational linguistics: a model "is better" if it performs better on a shared test set—according to various metrics. The relevant reference sets for MGs are already available as previously mentioned (SyntaxGym, §1). Another resource worth mentioning is the CoLA dataset—The Corpus of Linguistic Acceptability, Warstadt et al., 2018—which comprises approximately 10K sentences from various linguistics publications, annotated for (binary) grammaticality[14]. Similarly, the BLiMP dataset—a Benchmark of Linguistic Minimal Pairs for English, (Warstadt, Parrish, et al. 2020)—, includes 1K minimal pairs, where each pair is evaluated by native speakers for preference (A is preferred over B). Notice that the size of the dataset is not the most important feature: as previously mentioned in footnote 11, if a theory X can accurately predict the structure of a transitive sentence where a relative clause modifies one DP, as illustrated in (6).a, then it should also be capable of predicting an infinite number of similar sentences, where "similar sentences" means those sentences that follow the same instructions, rules, or principles, despite variations in lexical items that are irrelevant to the structure. Therefore, the quality of a dataset, will not be strictly related to its size in general, but to the number of truly different structural configurations included. Moreover, a dataset that includes subtle attested contrasts, for instance, in *subject islands* violations[15] (8), (Noam Chomsky 2008), will be more useful than one only including idealized ungrammatical cases (Huang 1982).

[14] An equivalent resource is available for Italian: the ItaCoLA dataset (Trotta et al. 2021).

(8) a. *[Of which car]$_i$ did [the driver _$_i$] cause a scandal?
b. [Of which car]$_i$ was [the driver _$_i$] awarded a prize?

It is not an accident that Chomsky and colleagues apply *the dust under the carpet principle* to islands by saying that "important judgments can be quite murky. Without a clear understanding of such data, competing analyses can be difficult to compare" (Noam Chomsky et al. 2023, 65). In fact, under Definition 2 and Definition 5, we have a practical way to assess theories that relegate "under the carpet" a considerable corpus of experiments on islands constraints sensitivity (Sprouse and Hornstein 2013): descriptive inadequacy.

### 2.3. *Objectives and Goals: an Intermediate Summary*

To summarize, we can draw two relevant conclusions: (i) without a fully explicit (formalized) (Minimalist) theory, progress will be limited (formalization issue); (ii) without focusing on a shared and complete empirical domain—namely, a common test set—we cannot effectively compare the descriptive and explanatory adequacy of different formalized theories (evaluation issue).

I believe that the partial fulfillment of these requirements has caused generative linguistics to lose its footing and become marginalized in the contemporary landscape, which is dominated by efficient computational models, sophisticated experimental approaches, and, crucially, powerful statistical inferential methods serving both domains.

Let us go back to the three perspectives, to draw an interim conclusion: First, the computational perspective appears to be leading in terms of observational adequacy, and possibly in terms of descriptive adequacy as well, unless we consider to what extent the "dust under the carpet principle" might fruitfully be used to exclude a long-tail or irrelevant idiosyncrasies. Moreover, while the computational perspective can address the learnability issue, it does so in a way that must be proven to be cognitively plausible (i.e., explanatory adequate). The theoretical perspective reminds us that restricting the empirical domain is necessary, due to the limitations imposed by our world knowledge, memory, and attention capacities. Unfortunately, by adopting this strategy, the most

---

[15] A *syntactic island* is a phrasal domain from which extraction is not possible. The term was coined by John Ross (Ross 1967) and has been applied to various domains, including the Subject domain —example (8).a—, or the Complex NP (Noun Phrase) domain —*Complex NP constraint* (Ross 1967, 127): "No element contained in a sentence dominated by a noun phrase with a lexical head noun may be moved out of that noun phrase by a transformation". This is illustrated by the example (i)—the NP is currently labeled DP, the sentence CP, and the offending gap linked to the extracted NP "the hat" (Ross 1967, 126):

(i) **The hat* which I believed [$_{DP}$ the claim [$_{CP}$ that Otto was wearing _$_{\text{the hat}}$ is red]]

refined formalizations of Minimalist Grammar capture only a limited number of idealized contrasts, in comparison to the extensive body of available psycholinguistic evidence. Finally, the experimental perspective underscores that factors such as memory and attention, along with other performance confounds, can be quantitatively measured. Consequently, there is no justification for unduly restricting our empirical data-cake. As a result, psycholinguistic theories are increasingly adopting machine-learning methods over explicit generalizations, even though the latter offer the significant advantage of intelligibility.

To convince ourselves that linguistics is not merely a rock-paper-scissors game, we must revisit some foundational issues and appreciate the advantage derived from an integration of these three different perspectives. The next section is devoted to the PoS hypothesis—a classic argument supporting the generative approach—and the notion of simplicity in structure-building operations, which is ostensibly a decisive factor in terms of learnability and evolvability. Indeed, I will argue that none of these issues can be adequately addressed from a singular perspective. Instead, all three viewpoints must be considered simultaneously to effectively tackle the Language Problem (Definition 1) and gain descriptive and explanatory adequacy.

## 3. Foundational Issues under Review

For a meaningful comparison of vLLMs and MGs that takes into account issues of formalization and evaluation, it is essential to revisit two foundational assumptions. These include the Poverty of Stimulus (PoS) hypothesis (§3.1) and the notion of simplicity in defining the core structure-building operation Merge (§3.2). The aim of this section is to underscore the importance of evaluating the impact of one specific formulation of structure-building operations against another. Ultimately, I will contend that although criticisms of vLLMs are warranted, they are frequently presented in an inappropriate or misguided manner.

### 3.1. The Poverty of Stimulus (PoS) Hypothesis: Learning from Positive Primary Linguistic Data

A cornerstone of generative linguistics is the PoS hypothesis, also known as Plato's problem (Noam Chomsky 1986). This hypothesis postulates the existence of innate knowledge to explain how children, under normal conditions and when sufficiently exposed to any natural language, can develop an adult-like linguistic knowledge—competence—that is not inferable from the limited exposure to the primary linguistic data available to them[16].

Let us consider both a stronger and a weaker version of the PoS hypothesis: under the stronger interpretation, no adult linguistic competence can be attained, regardless of the amount of primary linguistic data available as input. Under the weaker interpretation, an adult's linguistic competence can only be approximated after an 'unreasonable'—excessively large—exposure to primary linguistic data.

[16] It is important to emphasize that the term "poverty" can refer to both "quantity"—e.g., the number of words per year a child is exposed to—and "quality"—e.g., a consistent lexical and structural diversity in linguistic input. In this discussion, I will primarily focus on the quantitative aspect, under the somewhat overoptimistic, yet reasonable, assumption that sufficient and consistent diversity is present in common child-directed speech.

The stronger version is supported by classic results demonstrating the unlearnability (in the limit) of any recursive grammar, including the simplest regular ones (Gold 1967). These findings suggest that only trivial finite grammars are learnable when the primary linguistic input is limited to positive evidence only, thus excluding negative evidence, that is, information indicating that a sentence is ungrammatical. Negative evidence is systematically excluded as a significant source of information for children because it is not uniformly provided—it is not given in all contexts or to all children—and is often noisy or inconsistent (Guasti 2017).

This limitation to positive primary linguistic inputs serves as a safeguard against machine learning methods employing supervised learning approaches. Such methods necessitate training on both well-formed and ill-formed sentences and implement a rewarding policy based on the model's accuracy in classifying grammatical versus ungrammatical sentences received as input. Compared to the ecological conditions of the primary linguistic input, those approaches require a training process that is clearly cognitively implausible, then irrelevant to address the "logical problem of language acquisition"—"PoS" in other words. Machine learning approaches, however, adopted another method that easily circumvents the safeguard banning supervised learning: Elman (Elman 1990) proposed a training trick, dubbed *self-supervision*, which is still used by modern Transformers (Vaswani et al. 2017) to train the most powerful vLLMs. When these models are trained, the only task they need to perform is to predict, as accurately as possible, the next word/token in S. Given a context such as "please, open the …", the next expected token should be *something like* "door" or "window" and not like "is" or "John". "Something like", then, expresses a precise categorial generalization useful to provide a plausible token to successfully complete the next word prediction task. Simple Recurrent Neural networks (SRNs, Elman, 1990) perform nicely on this task simply using, during training, the incoming word to check the accuracy of their prediction—therefore the term "self-supervision". This solution is (a) easy to implement, (b) psycholinguistically and cognitively plausible (Cloze probability, Bloom & Fischler, 1980; Taylor, 1953, priming effects, Bock, 1986), (c) effective in capturing certain kinds of non-local dependencies (Elman 1993; Quinlan 2004). Finally, (d) it provides gradual predictions both off-line—an equivalent measure of the overall probability of generating a certain sequence S—and on-line—word-by-word, token-by-token preferences. It is, therefore, not surprising that this approach quickly gained popularity and reached a broad audience.

What is crucial to understand is that SRNs, among other capabilities, were able to learn certain non-local syntactic dependencies, such as subject-verb agreement, even when irrelevant lexical material (unrelated to the targeted agreement configuration) intervenes between the subject and the predicate. This is illustrated in (9)—as in (1):

(9) The dog [that the boys chase] grips/*grip the bone.

If trained SRNs prefer a third-person singular agreeing predicate over a non-agreeing one, regardless of the distance from the relevant subject, this indicates their observational adequacy in modeling such dependency. This preference suggests that SRNs implicitly prioritize grammatical structures over ungrammatical ones. Furthermore, this approach eliminates the distinction between idealized and externalized language (L) since it relies solely on observable (external) data to train and evaluate the model.

Many scholars have interpreted this as a significant challenge to the PoS hypothesis, suggesting that a tombstone may be placed over it. With a relatively simple "training program" and using plausible primary linguistic inputs, a substantial portion of our grammatical knowledge can be acquired. Specifically, SRN-like networks exhibit human-like preferences in critical linguistic distinctions without depending on the explicit tree structures ($T_S$) associated with sentences (Ss), which are obtained through abstract derivations ($D_S$). Although Elman himself clarified that these findings did not disprove the Poverty of Stimulus (PoS) hypothesis—arguing that the network architecture, training procedure, and learning algorithm could all be considered forms of "innate knowledge" (Bates et al. 1996)—the approach is generally perceived as a simpler, more elegant, and more appealing alternative to generative grammars and, subsequently, to Minimalism. This perspective has been reinforced by significant advancements: Long-Short Term Memory networks (LSTM, Hochreiter & Schmidhuber, 1997) and their successors draw on Elman's architectural and training insights and have introduced innovative solutions to address long-standing computational challenges, such as the vanishing gradient problem[17]. Recently, networks of the LSTM type have been shown to outperform the widely acclaimed Transformer models (such as those used to train GPT-X vLLMs) in handling complex tasks within very subtle syntactic islands domains (Wilcox, Futrell, and Levy 2023). To my knowledge, the paper by Wilcox et al. was the first to address various degrees of adequacy of these vLLMs in a typologically comprehensive manner on *Linguistic Inquiry,* which is probably the journal representing the most orthodox perspective on Minimalism[18]. Similar findings led some generative linguists to adopt a weaker version of PoS: while these models might possess the capability to discern significant linguistic contrasts, they require an excessive amount of data for learning, especially when compared to the relatively limited data received by the children (Katzir 2023). I think accepting a weak perspective on PoS is detrimental to the generative cause. However, this position offers a valuable opportunity to discuss several critical issues that the generative linguistic community has, perhaps neglectfully, adopted. Those lines of defense against vLLMs' descriptive and explanatory adequacy are generally linked to: (i) the excessive amount of data required to train these models (§3.1.1), (ii) the immense size of these models, which undermines their descriptive adequacy (§3.1.2), and (iii) the fact that language models lack an understanding of the meanings of the sentences they classify as grammatical (§3.1.3). I will address these arguments one by one in the following sections.

#### *3.1.1. Training Data Size Argument*

Many generative linguists may not yet be aware that, given a sufficiently large set of positive primary linguistic data—a size that can be reasonably quantified (Hsu and Chater 2010)—specific approaches, such as the Simplicity-Based framework (Hsu, Chater, and Vitányi 2013), allow for a relevant approximation of the *simplest* grammar, namely shortest in MDL terms (§2.1)[19]. The approach adopted effectively reduces both overgeneralization and undergeneralization errors, which are respectively the result of overly general—simpler, shorter—or overly specific—more complex, longer—grammatical rules, by prioritizing a reduction in description length for grammar formulation—akin to a computer program's length. Given that this framework can be logically implemented across various equivalent formalisms—e.g., mildly context-sensitive ones like Tree-Adjoining Grammars (Frank 1990; 2002), or MGs, (Stabler 2011; 2013)—, with no significant computational differences in terms of program length[20], it aligns with the goal of descriptive adequacy (Definition 2) as defined within the Minimalist Program. While finding the exact shortest grammar is a non-computable problem (M. Li and Vitányi 2008), a probabilistic approximation makes this goal attainable. Thus, if we adopt the weakest version of the PoS hypothesis, the size of the training set might remain the only argument against the viability of vLLMs from a generative linguistics perspective.

[17] In Recurrent Neural Networks that use standard backpropagation methods, we observe an exponential decrease of the error redistribution coefficient (i.e. the vanishing gradient) to be used to update the connection weights —i.e., learning— with, practically, no effect after few steps backward (S. Hochreiter et al. 2001).

[18] Several other relevant tests had already been documented in the computational linguistics literature (Chowdhury & Zamparelli, 2018; Linzen et al., 2016; Wilcox et al., 2018 a.o.) but were ignored by the generative linguistics community.

From this perspective, we have pretty accurate predictions of the average number of words children are exposed to as primary linguistic input during their early years: from 3 to 11 million words per year (Hart and Risley 1992). A clear trade-off must then be considered: the complexity of a rule directly impacts the amount of data needed to integrate such a rule into our grammatical knowledge[21]. So, approximation becomes a tricky issue here: is it sufficient to obtain a reasonably compact grammar that captures 99.9% of the data in a test set? I believe the answer depends on the dataset. If the test set accurately reflects a comprehensive range of linguistic configurations, especially those rarely found in natural speech but documented by Stage of Acquisition (SoA), achieving this would be a significant accomplishment. However, if the test set is limited to statistically frequent or naturally occurring constructions, the value of such a result is comparatively limited. The key challenge lies in testing our internal (I-language) competence using an external (E-language) dataset—a corpus specifically designed to challenge theoretical assumptions.

Just to remain on “murky judgments”, let us again consider the subextraction of a wh- item from a direct object, (10).a, and a context where, to avoid wh- subextraction from a complex NP (strong island), we rely on alternative structural solutions, like (10).b. Can our learning procedure, which is essential for achieving explanatory adequacy, infer from very limited data that the avoidance of the application of a “rule”—in (10).b—that applies to (10).a, in fact, indicates a prohibition against configurations like (10).b'?

---

[19] This, in fact, does not constitute the first challenge of Gold’s unlearnability results (cf. Clark & Lappin, 2010). The learnability of general rules (e.g., the 'pair' and 'append' operations on strings, which may encompass some key insights underlying the minimalist 'Merge') is also discussed in Y. Yang & Piantadosi (2022).

[20] This is the *Invariance theorem* (M. Li and Vitányi 2008). In their experiment, Hsu & Chater (2010) employ a probabilistic context-free grammar, the adequacy of which we might reasonably question. The same experiment could be replicated using probabilistic MGs (J. Li and Hale 2019), potentially addressing phenomena more complex than the 19 contraction rules investigated in the original paper by Hsu & Chater.

[21] Another way to formulate the trade-off between regular (simpler) rules and exceptions is the *Tolerance Principle* (C. D. Yang 2016).

(10) a. Who did you see a painting of _?
b. Did you see a painting representing whom exactly?
b'. *Who did you see a painting that was representing _ exactly?

Obviously, (10).b' cannot be found in any naturalistic positive primary linguistic input. The only method to evaluate the theory's ability to make relevant generalizations is by incorporating examples like (10).b' into the shared test set. This means relying on any subtle empirical evidence available, crucially including explicit and controlled judgments in our test set. Although such examples may comprise only 0.1% of the test data, their ability to induce model prediction failures is telling. This would indicate that the seemingly accurate generalizations regarding (10).a-b are misleading. This conclusion stems from the realization that these generalizations rely on a non-uniform—or "genuine"—interpretation of wh- dependencies, failing to account for the ungrammaticality of (10).b', as well as structurally equivalent infinite lexical variations of it.

From this viewpoint, arguing against the PoS hypothesis based solely on the size of the training set might not only be unnecessary but also counterproductive. A more compelling argument lies in the irrelevance of the volume of positive data received during training if a language model fails to generalize critical linguistic facts, such as the ungrammaticality of specific island violations, in line with robust native speakers' judgments. Therefore, the careful selection of a shared test set, one that includes ungrammatical sentences, becomes crucial. Such a dataset is essential to demonstrate the enduring relevance of the PoS argument and to show that the grammar inferred by the model deviates from human-like performance.

Consider now the classic debate on the unlearnability of auxiliary-subject inversion in English yes-no questions, addressed by Crain and Nakayama's experiment (Crain and Nakayama 1987). This experiment can be adapted to test various vLLMs to see if their strategy for forming 'yes-no questions' aligns with the structural solutions observed in children aged 3-5 years. Crain and Nakayama employed a straightforward elicitation task: providing a context targeting a character X, they prompted children to ask X a question like, 'Ask X if the boy who *is* watching Mickey Mouse *is* happy.' Despite various errors, children consistently avoided incorrect inversion of the embedded auxiliary with the matrix subject—i.e. sentences like "**is* the boy who $_{is}$ watching Mickey Mouse *is* happy?" are never produced. This is a robust finding that persisted across tense changes—e.g., '...*is*/*was*... *was*/*is* happy'—and the substitution of an auxiliary with a modal—e.g., *'is*/*can*... *is* happy'. Interestingly, ChatGPT (based on the GPT-3.5 version as of May 2023) managed to correctly perform the matrix subject-auxiliary inversion without difficulty. This experiment might suggest that traditional arguments must be refined and that more evidence is needed to delve deeper into the PoS debate (§3.1.2).

Generative linguists exploring the implications of training language models with realistically plausible data volumes should examine the findings from a recent shared task, BabyLM (Warstadt et al. 2023) before fully embracing arguments based on training size. In summary, BERT-based optimized models (ELC-BERT, Charpentier & Samuel, 2023) achieved an accuracy of approximately 0.85 (with 1 representing perfect accuracy) in distinguishing minimal pairs from the BLiMP dataset (Warstadt, Parrish, et al. 2020) when trained on a plausible corpus of 100 million tokens. Their accuracy decreases to 0.80 when the training size is reduced to 10 million tokens. To be precise, achieving the mentioned accuracy required 450 epochs, meaning each sentence in the 100M token dataset was presented to the model 450 times[22]. For the 10M token training, this number escalated to 4000 epochs, highlighting a significant efficiency concern. A child is a real "few-shots learner" (Brown et al. 2020) in this respect. Further, two additional outcomes underscore the linguistic limitations of these models: firstly, the ELC-BERT system (Charpentier & Samuel, 2023), despite winning the BabyLM competition, only achieved an accuracy of around 0.47 on the Mixed Signals Generalization Set (MSGS, Warstadt, Zhang, et al., 2020)—a benchmark consisting of ambiguous binary classifications designed to test a model's preference for hierarchical versus surface-level generalizations—and 0.59 on CoLA (Warstadt, Singh, and Bowman 2018). Secondly, while accuracy improves with more training, the models' alignment with online metrics, such as reading time, deteriorates (Steuer, Mosbach, and Klakow 2023) suggesting a diminishing return on linguistic relevance with increased training. Putting all this evidence together, I conclude that the stronger version of the PoS argument remains unchallenged by pre-trained vLLMs currently tested.

Bear in mind that this conclusion is crucially supported by on-line performance data experimentally collected on subtle linguistic contrasts.

#### *3.1.2. Descriptive Adequacy Argument and Data Problems*

Connectionists (Rumelhart, McClelland, and PDP Research Group 1999) heralded the transition from the traditional competence/performance distinction to a unified processing approach with considerable pride. The advantages appeared to be substantial: connectionist models can predict a wide range of data types, including (i) implicit and explicit, (ii) categorical and gradual, (iii) off-line and on-line, (iv) ecological and controlled. This significantly raises the benchmark for observational adequacy: given their capacity to incorporate a broader spectrum of data, connectionist theories have the potential to achieve greater observational adequacy. Once more, the generative linguistic safeguard is provided by Definition 2: the quality of data—encompassing generalization aspects beyond simple quantity, as discussed in sections §2.2.5 and §3.1.1—and the principle that fewer instructions lead to better descriptive adequacy. This principle highlights a critical vulnerability of these models: their substantial size. The dimension of any sub-symbolic model can be expressed in terms of the number of parameters: this has nothing to do with linguistic parameters (Noam Chomsky 1981), but it is a useful approximation of the model's learning dimensions. That is, a 10-parameter model has 10 dimensions to represent the problem. In addition to these dimensions, one must consider the length of the computer program required to utilize these parameters. However, due to the Invariance Theorem—footnote 20—, this factor is typically negligible in most scenarios. In the context of modern vLLMs, GPT-3 is known to have 175 billion parameters. While the exact size of GPT-4 is not publicly disclosed (rumors indicate ~1.8 trillion parameters), the scaling hypothesis (Kaplan et al. 2020) implies a significant increase in parameters is necessary for improved performance. This represents a degree of complexity far surpassing that of any MG formalization[23].

---

[22] The situation is slightly more complex than it appears: a more accurate method for calculating data exposure should take into account the training steps, approximately 31k, and the batch size, about 8k, as reported in the appendix of the original paper for the 100M challenge submission. However, exploring the complexities of these concepts and the optimization steps falls outside the scope of this work.

[23] Including into a MG formalization both the lexicon and its features, using annotated Universal Dependencies treebanks (Nivre et al. 2017) to consider the most relevant kinds of lexical and syntactic ambiguity, we can estimate a lexicon size from about 13K distinct entries in English to about 33K in Turkish

Therefore, even when considering the Simplicity-Based account (Hsu, Chater, and Vitányi 2013), MGs are posited to surpass these vLLMs in the Descriptive Adequacy competition. This is obviously a wrong conclusion primarily because the full test set MGs can account for is significantly smaller than that addressed by vLLMs. Standard MG implementations I am aware of (Ginsburg and Fong 2019) are limited to categorical grammaticality judgments, making them suitable for only a narrow range of predictions on naturalistic or gradual linguistic data. Incorporating statistical considerations (Hale 2011; 2001) is necessary to extend MGs applicability towards performance data. The situation marginally improves with the adoption of more processing-friendly MG formalisms (Chesi 2021) or parsing-based metrics (De Santo 2019; Graf, Monette, and Zhang 2017; J. Li and Hale 2019; Stabler 2013). These adaptations allow MG-based models to handle a broader variety of linguistic data types (§2.2.1-§2.2.4). Despite their enhanced ability to make linguistically informed predictions on minimal contrasts—a capability only recently tested against vLLMs (Hu et al. 2020)—the overall alignment of MG-based models with naturalistic data remains lower if compared to advanced surprisal measures (Futrell, Gibson, and Levy 2020; Futrell and Levy 2017; Hale 2016; Levy 2008) based on robust statistical models[24].

Summarizing, vLLMs demonstrate remarkable robustness in harnessing diverse data sources obtained through sophisticated experimental methodologies, readily incorporating various empirical linguistic dimensions without hesitation. In contrast, formal models efficiently handle binary judgments but falter in capturing continuous, nuanced predictions. These theoretical models necessarily resort to "the dust under the carpet principle", ultimately marginalizing themselves from significant benchmark competitions. This is particularly evident in tests featuring relevant contrasts that do not pose any problem for a theory fragment taken in isolation but pose significant challenges to the overall theoretical framework consistency. Therefore, in challenging bulk testing scenarios, vLLMs exhibit a clear empirical advantage.

*Data Dust: What Empirical Evidence Should We Ignore?*

We are left with the uninterpretability of the implicit representations of these models, but this is, again, a hardly sustainable criticism. If a model's description is shorter and thus descriptively more adequate, it ought to be preferred, as it is expected to capture essential generalizations despite their not being immediately intelligible. This dilemma aligns with the issues raised in Piantadosi's paper, which argues that if two equations yield identical predictions with equal computational costs

---

(Chesi 2023). Without considering efficient morphological decomposition (Kobele 2023), a full MG grammar is unlikely to require more than 100K parameter-equivalent dimensions. By approximation, we can equate this to the number of bits, as mentioned in footnote 7, which results in a size increase of at worst one order of magnitude—the compressed Python code, which includes the extracted English lexicon and is freely available at github.com/cristianochesi/e-MGs, has a size of merely 177KB. This size difference is substantial, spanning several orders of magnitude, especially when compared to the parameter scale of GPT models.

[24] A "surprisal-like" measure is a value derived from information-theory metrics, indicating the unexpectedness of a word given its preceding context. The higher the surprisal value, the less likely the word is perceived as a natural continuation of the sentence. Such measures are utilized to predict processing difficulties in real-time reading or listening scenarios, with greater surprisal indicating higher expected difficulty. Additionally, surprisal values can be used to predict the outcomes of offline tasks: by calculating the cumulative surprisal for each word in a sentence, one can assess the overall unexpectedness of such sentence. For example, in forced-choice tasks, the sentence with a lower total surprisal score is generally predicted to be more easily accepted or understood.

(in terms of steps taken or memory/dimensions utilized), they should be deemed equivalent in terms of descriptive adequacy.

In my view, however, there exists a compelling argument for favoring MG-based approaches over vLLMs on similar grounds. If vLLMs are indeed capable of encapsulating all relevant syntactic and semantic generalizations after relatively modest data exposure (between 10 million to 100 million tokens), with the remainder training corpus associated to “commonsense knowledge” (Zhang et al. 2020), then we should narrow our focus to the modest set of parameters encoding structural generalizations, fruitfully adopting the “dust under the carpet” principle. By doing so, we can undertake a precise comparison with MGs. This approach allows us to evaluate the efficiency and efficacy of each model in capturing the core linguistic (syntactic) phenomena, setting a more focused and fair ground for comparison. The rationale behind this comparison is that core syntactic properties ought to account for most data compression, leaving the idiosyncrasies to consume the bulk of the parameters in a vLLM. We can then attempt to define which types of data may be excluded from our test set by adopting a “Data Dust” principle:

**Definition 6.** *Data Dust*
Irrelevant data for linguistic theorizing are those that force an increase in theory size without yielding any new generalizations

In more explicit terms, if a theory X for language L predicts a novel contrast $C_n$ (e.g., a minimal sentence pair) without incrementing its size, then X already encompasses the necessary generalization. Thus, $C_n$ becomes a valid candidate for inclusion in a test set for language L. Conversely, if no linguistic theory performs a correct prediction on $C_n$ without increasing in size, and there are no other contrasts $C_m$ that this size increase captures, then $C_n$ qualifies as “data dust”. $C_n$ can then be temporarily relegated “under the carpet” until a new theory Y emerges that can predict it without increasing its size.

Measuring exactly the tolerable compression rate necessary to classify a contrast as relevant or not is out of the scope of this paper—this might be crucial for distinguishing linguistic competence from world knowledge—, but it seems to me clear that this perspective challenges the sustainability of probabilistic learning approaches as proposed, for example, by Hsu et al. (2013). Specifically, it posits that learners must cope with complex NP island constraints even in the absence of direct evidence for wh-item sub-extraction from a relative clause, a point elaborated in §3.1.1. From this viewpoint, the key is not merely accumulating a vast dataset but compiling a targeted collection of examples (possibly including minor lexical variations) that test the model's handling of configurations unlikely to be found in naturalistic data. This approach is exemplified by the generation of pattern-based items in the AcCompl-IT dataset (Brunato et al. 2020), which are evaluated by human judges and all produce consistent responses (based on a 7-point Likert scale). Such phenomena include cases like illegally filled gaps (11) or unlicensed negative polarity items (12).a, areas where the performance of GPT-based models remains imperfect—they randomly accept gaps filled illegally, example (11) ending with "-lo", and they license negative polarity items also in the absence of an appropriate negative licensor, example (12).a ending with "mai"[25].

(11) {Che cosa | quale problema}$_i$ lo studente dovrebbe descriver(e) {_$_i$|*-lo$_i$ }?
{that what | which problem}$_i$ the student must describe {_$_i$|*it$_i$}
{*what | which problem*}$_i$ *must the student describe* {_$_i$ | **it*$_i$}

(12) a. Maria si aspetta che qualcuno possa avere {già | *mai}
M. himself expects that someone could have {already | *ever}

finito questo esercizio
completed this exercise

b. Nessuno si aspetta che qualcuno possa avere {già | mai}
Nobody himself expects that anyone could have {already | ever}

finito questo esercizio
completed this exercise

{*M.* | *Nobody*} *expects* {*someone* | *anyone*} *to have* {*already/ever*} *finished this exercise.*

I remind the reader that vLLMs are designed primarily for answering questions and executing various NLP tasks, rather than explicitly modeling human grammatical competence. These models, including GPT-4, represent a substantial advance in encapsulating a broad ontology—a detailed and structured body of knowledge that enables them to tackle challenges like the Winograd schema (Levesque 2014) and excel in numerous knowledge-based assessments (OpenAI 2023). Within this broad spectrum of capabilities, grammatical knowledge is merely one aspect under evaluation.

The performance of these models in dealing with complex grammatical distinctions, despite being trained for a general next-word prediction task, might seem remarkable. However, this capability is not entirely genuine. Supervised fine-tuning is crucial for transforming these systems into effective "few-shot learners" (Brown et al. 2020). The nature and quality of this "fine-tuning"—essentially a form of supervised learning—remain opaque. Consequently, when considering these models as "grammatical theories" in the vein proposed by Piantadosi, their relevance becomes questionable. If the fine-tuning process leverages comprehensive datasets such as CoLA, it could render the learning challenge both biased and unreasonable, thereby diminishing the models' relevance as cognitive models of language. This concern underscores my skepticism toward assessing linguistic competence by requesting grammatical judgments from ChatGPT.

*On the Separation of PF from LF: A Challenge for Descriptive Adequacy*

To conclude, returning to the T-model as outlined in (5), and the proposed division of labor between core syntax, PF, and LF, it is now crucial to reevaluate the extent to which this minimalist perspective provides a logical advantage over the integrated approach to processing, semantic, and syntactic knowledge offered by vLLMs.

[25] In Italian, "mai" (ever) is a negative polarity item which must be licensed by a C-commanding—see footnote 9—negation like "nessuno" (nobody), as in the example (12).a. Replacing "nessuno" with "Maria", which is not a negative item, "mai" remains unlicensed. This explains the ungrammaticality of "Maria si aspetta che qualcuno possa avere *mai finito questo esercizio" (M. expects someone to have ever finished this exercise).

A critical empirical challenge for the T-model, which cannot easily be swept under the carpet, concerns the interpretation of quantifiers and their scope. It has been proposed that Quantifier Raising (QR, May, 1985) should be considered "the dust under the carpet", namely a matter for the LF component. However, under certain very fruitful analyses, this operation is necessary to predict *extrapositions*, an operation that impacts on linear order—namely on the overt realization of S—as exemplified by the contrast (13).b-b' below:

(13) a. I saw [{a (very good) / the (best)} picture [of the museum]] yesterday.
b. I saw [a (very good) picture $_ _i$ ] yesterday [of the museum]$_i$.
b'. ??I saw [the (best) picture $_ _i$ ] yesterday [of the museum]$_i$.

In a nutshell, the prepositional phrase [of the museum] can be "extraposed" from its host, [a (very good) picture], in (13).b, but crucially not in (13).b', due to the quantificational status of the determiner "a" as opposed to "the" (Baltin 2017; Fox and Nissenbaum 1999). If QR, which is supposed to happen optionally at LF, has an effect on linear order, which is supposed to be relevant only at PF, either we remove the optionality of QR from the theory (Beghelli and Stowell 1997)—but then how can we account for the optionality of extraposition? This is necessary, as suggested by the grammaticality of both (13).a and (13).b—or, otherwise, there is no clear way to relate LF phenomena with PF effects. Consequently, a theory that explicitly addresses these constraints, and integrates them within a unified structural framework without an increase in size (Definition 6) would offer a better descriptive adequacy than the orthodox T-model. Unfortunately, to the best of my knowledge, neither MGs nor vLLMs have been tested against these specific contrasts.

#### *3.1.3. The Misunderstanding Argument*

Despite their impressive ability to answer specific questions accurately, it is often highlighted that vLLMs do not truly "understand" complex, well-formed sentences and can sometimes process incorrect ones—or "impossible", in the sense described by Moro (Moro 2023; Moro, Greco, and Cappa 2023). This criticism must be handled with care for precise reasons that are worth exploring in some detail.

Roni Katzir, for instance, responding to Piantadosi's paper (Katzir 2023), tried to elicit from ChatGPT paraphrases that require the correct structural analysis, as in the case of (6).a here repeated for convenience:

(6) a. (I saw) [a mouse [that [a cat [that [a dog bit]] chased]] ran away].

Katzir showed that some versions of the GPT model struggled to identify who was chasing whom. In my opinion, this is the only crucial test to determine whether a model's understanding of a sentence aligns with the assumed structural representation T or not. It is crucial to remember, however, that a comparison with human performance on these intricate test sentences remains necessary (§2.2.5). Another sound approach is to rely solely on the accuracy, stability, and convergence of judgments (Dentella, Günther, and Leivada 2023).

Alternative strategies to prove the misrepresentation of T that are based on (i) illogical answers, (ii) incorrect metalinguistic explanations, or (iii) answers to ill-formed inputs are less effective. As far as

(i) is concerned, asking logical or ethical reasoning (Noam Chomsky, Roberts, and Watumull 2023) can be useful to assess the social danger related to vLLMs, but does not provide any compelling evidence about the linguistic knowledge expressed by these models. This follows under the reasonable assumption that linguistic competence and other high cognitive functions (including reasoning, theory of mind, etc.) are independent segregated modules, in Fodor's sense (Fodor 1983), as explicitly highlighted also in Piantadosi's paper. As for (ii), asking for metalinguistic explanations (non-intuitive judgments) and then criticizing imprecise usage of morphosyntactic terms is extremely unfair: apart from linguistically educated native speakers, for instance, no one have a clear understanding of why the nominative case is generally associated with the grammatical subject in languages like English. As for (iii), asking ill-formed questions and collecting reasonable or unreasonable answers are equally weak approaches: standard Minimalism does not make any explicit assumption on this performance side. In other words, native speakers might indeed be capable of answering questions by identifying and correcting a subject-verb agreement error. However, understanding how this correction process works falls outside the computational scope of any MG as currently conceived. Similarly, it is well-documented that humans can easily misinterpret ill-formed, nonsensical questions (Bever and Townsend 2001), yet no MG tackles this issue. Following the set-theoretic tradition, Definition 1 (Language Problem), incorporates both positive and negative criteria—"*all and only* the sentences Ss belonging to language L". The positive criterion ensures the model can capture any sentence in language L, while the negative criterion excludes sentences not belonging to L. The negative restriction seems to be overlooked by vLLMs, which tends to be 'collaborative' with ill-formed inputs. Criticisms built on these arguments would just reinforce the idea that those models are more descriptively adequate (robust) than any MG since they also model how we could recover from ill-formed inputs by re-analysis. Proving that the recovery/re-analysis strategy is adequate or not is a matter for empirical investigation, requiring the collection of experimental data. It is important to stress again that mainstream MGs are unable to provide any such strategy.

#### *3.1.4. Taking Stock of the PoS Hypothesis*

To wrap up, in section §3.1, I argued that the PoS hypothesis remains unchallenged by results from current vLLMs, even though classic arguments might require refinement—this is the case, for instance, with auxiliary subject inversion in English polar questions, which appears to be too easy a task for vLLMs. I have also claimed that standard criticisms of these models, which are based on three principal arguments—training data size, model size, and the misunderstanding argument—are at best irrelevant and at worst ill-posed. A more substantial criticism relies on models' ability to encompass infinite generalizations without increasing in size. Given that neither vLLMs nor MGs have been evaluated on this specific criterion, their descriptive and explanatory adequacy is yet to be determined. So far, we can conclude that vLLMs are observationally more adequate but lack explanatory adequacy. MGs are more compact and intelligible. This suggests that, when evaluated with a comparable test set, their descriptive adequacy could potentially be superior.

### *3.2. The Simplicity Mantra: Merge, Linear Order, and Cross-Linguistic Variation*

To conclude this paper, I aim to discuss two remaining issues related to the core assumptions of Minimalism: the simplicity of structure-building operations and the external constraints influencing these operations—third factors (Noam Chomsky et al. 2023). I think that these two fundamental issues contributed to the perception of the Minimalist program as an elusive framework.

#### *3.2.1. Simplistic Structure-Building Operations*

A little personal anecdote to introduce the issue: in 2002, I was preparing a seminar with Klaus Ables for Noam Chomsky and Danny Fox's class at MIT on "foundational issues". The topic assigned to us was "Vision and Language". The reference list we received notably included Marr's monumental work on Vision, unsurprisingly endorsing the generative grammar approach (Marr 1982, 28–29). During the preparation for our seminar, I had the opportunity for a brief interview with Tomaso Poggio (Marr and Poggio 1976; Riesenhuber and Poggio 1999). Then I inquired about the extent to which a Merge-like operation, described as "the simplest possible structure building operation, namely a set-formation operation", could be integrated as a computational component in vision modeling. Poggio's unequivocal reply is etched in my memory: "This does not make much sense to me, it's too simplistic". This could likely have been seen as a quick dismissal of an overeager PhD student's question, but it persuaded me that, despite the fundamental necessity of an operation to create structure—which undoubtedly must be "as simple as possible"—the Merge operation might be overly simplistic.

Without beating around the bush, an operation that overgenerates systematically is, computationally speaking, useless. It is easy to show that the unconstraint Merge operation used in (3) can produce whatever ungrammatical sentence we want—e.g., Merge("scolds", Merge("Alice", "Bill")) = {scolds {Alice, Bill}}[26].

Constraints come with a high cost, and due to Definition 2, a filter that excludes already generated unwanted structures will be discarded in favor of simpler (shorter, in description length terms) solutions. Although historically, a strategy involving filters was explored (Noam Chomsky and Lasnik 1977), it is more convenient to pursue a more descriptively adequate solution that simply better constrains the derivation, thereby reducing unnecessary computation (Frampton and Gutmann 2002).

In their recent crystallization of Merge operation, Chomsky and colleagues (Noam Chomsky et al. 2023) rely on Stabler & Collins formalism (Collins and Stabler 2016) and defend the set-formation idea behind this operation, as formulated in (3), under the simplicity and evolvability lens: the core operation to originate phrase structure must have been simple enough to evolve from a minimal genetic modification in our ancestral DNA. This line of reasoning does not extend to essential constraints like *labeling*, which involves selecting only the relevant information from the merged set that is useful for subsequent operations (Bošković 2016; Noam Chomsky 2013; Rizzi 2016). Roughly speaking, labeling is necessary, for instance, to distinguish a verb phrase from a noun phrase without inspecting the content of the phrase built so far.

If labeling is not "part of Merge" or something that must be learned, but rather a "third factor"—a "natural law" characterized by efficiency or "minimal search"—then why haven't we similarly recognized another widespread natural law like *incrementality*? Incrementality profoundly influences our language performance, both in comprehension (Bever 1970) and production (Bock

[26] In this derivation "{Alice, Bill}" would be predicted to be a constituent, but this violates any reasonable syntactic test—e.g., wh- substitution of {Alice, Bill} "constituent". Moreover, the predicted linear orders for this sentence would be <scolds, A., B.>, <scolds, B., A.>, <A., B., scolds>, or <B., A., scolds>, none of which are attested, as base-generated, in standard English.

1986). Given that phrase-structure building inherently progresses in (abstract) time, affecting every derivational approach crucially, it remains a mystery to me why this has not been recognized as a significant "law of nature"[27].

To appreciate this critical point let us return to the distinction between competence and performance. It logically follows that any performance-related task shall rely on one unique competence. Those tasks crucially include both comprehension and production—or parsing/recognition and generation, in computational terms, as overtly stated in the Language Problem (Definition 1). The fact that we need to draw upon the same competence theory is evident in the case of the lexicon: duplicating lexical entries to suit separate tasks for parsing and generation is highly inefficient. Yet, why has not this principle been equally applied to structure building? Merge operation, as conceived in (3), is incompatible with incremental parsing. This incompatibility arises because the order in which words are processed during sentence comprehension is the exact reverse—i.e., i. {Alice, scolds}, ii. {Alice, {scolds, Bill}}— of the order in which the structure (4) is derived according to (3)—i.e., i.{scolds, Bill} ii. {Alice, {scolds, Bill}}. As a consequence, a parsing/recognition algorithm must "undo" Merge and Move, trying to guess the non-deterministic operations that might have resulted in specific word orders (Stabler 2013).

It is worth recalling that the original concept of perfection was aimed at finding an ideal solution for interface conditions (§2.1). Now, under the multiple spell-out approach, we have come to envision our core language faculty as a generative process that directs elements to PF and LF. Elements that, ironically, cannot be readily pronounced or (completely) interpreted!

This problem is easily illustrated for PF by example (6).a, which is here repeated with an indication of the timing of the spell-out points associated with the CP phases:

(6) a. [$_{ph\ 4}$ I saw a dog [$_{ph\ 3}$ that bit a cat [$_{ph\ 2}$ that chased a mouse [$_{ph\ 1}$ that ran away]]]].

Focusing solely on the complementizer layer and not taking other potential phasal domains into account, Phase 1—'(mouse) that ran away'—will be sent to PF before Phase 2—'(cat) that chases a mouse [$_{ph\ 1\ (already\ spelled\text{-}out)}$]'. Such a derivation is logically possible, but it is completely non-sensical from a performance perspective. For the articulation of the complete sentence, PF should logically wait until the highest phase is complete before proceeding, but this approach is both counterintuitive and logically flawed. It is counterintuitive because speakers frequently commence utterances without having fully planned subsequent modifications, such as the addition of relative clauses towards the end of a sentence. It is logically flawed because it disregards the phenomenon of unbounded right recursion illustrated in (6).a, which does not present a processing challenge. Such recursion demonstrates that the core syntactic engine inherently supports incrementality—producing sentences piece by piece. Unfortunately, this is not achievable with any mainstream MG that relies solely on Merge, as discussed in §2.1.

[27] Kayne explored the relevance of timing in his influential work (Kayne 1994), but to the best of my knowledge these considerations have not been pursued further.

One might be misguided by the fact that incrementality seems to correspond to linear order, but this is an illusory perspective. Maintaining the premise that Merge merely constructs structure, as defined by (3), one might assume that:

i. Merge is binary because of time ("third factor"): the incoming token ($\beta$) must be merged within the already formed structure ($\alpha$);
ii. Merge simply produces hierarchical structures, that is MERGE($\alpha$, $\beta$) = [$\alpha$ [$\beta$]]

I fail to see how these two points contribute to an increased complexity of the Merge operation beyond that which is present in (3). However, I do recognize several benefits arising from this shift. The incorporation of incrementality considerations (Phillips 1996) represents a significant enhancement concerning interface conditions. In fact, this should be the null hypothesis (Momma and Phillips 2018) as it directly facilitates the delivery of syntactic objects that can be pronounced and interpreted incrementally. Consequently, it becomes possible to manipulate features that regulate nesting and movement [28], thereby obviating the need for extraneous labeling considerations: the "label" becomes the item that selects/expects another item as its complement. The alignment of this selection/expectation concept with the foundational principles of Elman's SRN and subsequent research is, in my view, remarkable. Determining whether specific artificial neural network architectures provide equivalent algorithmic solutions (in the sense proposed by Marr) for the computational and algorithmic predictions derived from this novel interpretation of Merge remains a question for empirical investigation. For example, the constraints associated with the application of this selection/expectation-based operation to (strong) islands (Bianchi and Chesi 2014) should emerge not from learned behavior but as a direct consequence of formalizing a structure-building operation that incorporates considerations on both nesting and incrementality. In this context, networks resembling SRN or LSTM appear more adept at reflecting the concept of incrementality than those based on attention mechanisms. This observation aligns with empirical findings (Wilcox, Futrell, and Levy 2023).

As discussed in §2.2.5, Stabler undertook the task of addressing numerous omissions in the original definition of Merge to delineate a well-defined structure-building operation capable of generating meaningful phrase structures. The introduction of feature checking, for instance, and the induction of linear order, may have been met with skepticism by some scholars. Nonetheless, these extensions were essential in rendering a particular variant of MG empirically testable. Other intuitions might well be formally articulated and empirically evaluated: If Merge is posited to be independent of feature checking, thus requiring a Labeling algorithm (Noam Chomsky 2013; 2015), then it becomes imperative to demonstrate that derivations pertaining to the same test set can be accounted for in potentially more efficient and descriptive adequate ways than those offered by alternative feature-checking formulations. This evidence is currently missing especially in the case of (strong) islands.

[28] E.g., Complementation: MERGE($\alpha_{=x}$, ${}_{x}\beta$) = [$\alpha_{\cancel{=x}}$ [${}_{\cancel{x}}\beta$]] or MERGE(${}_{x}\alpha$, $\beta_{=x}$) = [[${}_{\cancel{x}}\alpha$] $\beta_{\cancel{=x}}$]
Free adjunction: MERGE($\alpha_{=x}$, ${}_{y}\beta$) = [[${}_{y}\beta$] $\alpha_{=x}$]
Movement: MERGE([[${}_{y}\beta$]$_{i}$ … $\alpha_{\cancel{=x}}$ [${}_{\cancel{x}}\gamma_{=Y}$]], [${}_{y}\cancel{\beta}$]$_{i}$) = [[${}_{y}\beta$]$_{i}$ … $\alpha_{\cancel{=x}}$ [${}_{\cancel{x}}\gamma_{\cancel{=Y}}$ [${}_{\cancel{y}}\cancel{\beta}$]$_{i}$]]

### *3.2.2. Generative Parameters and Word Order Variation*

After the Pisa lectures (Noam Chomsky 1981), considerable research effort has been directed toward identifying a comprehensive list of parameters and organizing them into a coherent hierarchy (Baker 2001). This endeavor aims to render the problem of learnability more manageable. In this domain, I perceive the most significant advancements within generative linguistics: on the one hand, there has been a significant extension of the empirical basis, driven by radical generalizations regarding the restrictions of functional sequences (Cartographic approach). On the other hand, there has been a refinement in data elicitation methods that allows for an effective investigation into micro-parameterization, alongside the development of sophisticated mathematical methods for calculating phylogenetic distances (Gianollo, Guardiano, and Longobardi 2008; Guardiano et al. 2020; Guardiano and Longobardi 2016). A coherent picture is emerging in which parameters are not considered inherent components of Universal Grammar. Instead, they are viewed as options that remain underspecified until activated by specific selections of (lexical) feature bundles (Roberts 2019). Various resources sprouted from the cross-linguistic perspective, including atlases of different kinds. Notably, these include the World Atlas of Language Structures, WALS (Dryer and Haspelmath 2022), and the Syntactic Structures of the World's Languages, SSLW (Collins, Kayne, and Koopman 2009). Once again, the most straightforward method of engaging with heterogeneous resources involves depending on the least ambiguous evidence they offer. For example, this can include preferences in word order expressed in terms of specific categories. An illustrative example is provided by Greenberg’s Universal 20 (GU20), which focuses on word order within the extended nominal domain—DP. An original analysis emerged from Cinque’s derivation of GU20 (Cinque 2005; Roberts 2017). Without digging into technical details, the attested and unattested word orders of four pertinent categories within the DP domain can be predicted by positing a universal hierarchical ordering of these categories, specifically [Dem [Num [Adj [NP]]]], plus a set of constraint on (head) movement. Cinque assumes Kayne’s *Linear Correspondence Axiom* (LCA) to justify some of the constraints imposed on movement (Kayne 1994), while others not only consider this approach redundant but also dismiss LCA on the basis of an orthodox conception of Merge which expresses no order between the elements merged. According to this second view, if Merge operates on Adj and NP, then both possible linear orders, <Adj, NP> and <NP, Adj>, could be observed across different languages (Abels and Neeleman 2012).

Both the approach relying on Kayne’s LCA and the critique offered by Abels and Neeleman present plausible arguments that warrant thorough evaluation. While Abels and Neeleman’s critique raises significant questions about the formal legitimacy of LCA, offering what appears to be a more elegant solution to the generalizations of GU20, an ultimate comparison of these proposals—considering simplicity, descriptive adequacy, and explanatory power—must be achieved. Moreover, an examination of the extensive empirical evidence assembled by Cinque, particularly with reference to data detailed in a dedicated section of the SSWL, uncovers instances of data idealization. This suggests that sweeping “the dust under the carpet” remains a necessary principle for navigating through noisy data while still benefiting from coarse-grained categorial idealizations that can be later refined. While data idealization can accelerate early-stage research, the increasing precision of modern linguistic analysis necessitates a robust methodological approach. This approach should mirror established experimental practices and emphasize the creation of openly accessible resources. Such resources are essential for rigorous testing of any fully formalized theory.

In conclusion, formalizing fragments or assessing specific phenomena is essential for progress; however, in the absence of a unified framework in which to integrate coherently these fragments, the resulting overview appears as a Cubist patchwork. It is noteworthy that every complex discipline encounters similar crossroads: the Standard Model in quantum physics, for instance, is often criticized as a patchwork (Oerter 2006). Despite widespread dissatisfaction with this, it remains effective. It offers, to date, the most concise description of reality available, predicting phenomena with the highest accuracy possible. This includes phenomena that are rarely observable under natural conditions but can be artificially reproduced, such as those observed in the Large Hadron Collider (LHD, Evans, 2007). By controlling the size, speed, and position of a few particles, researchers can capture better pictures of the collision events.

It appears to me that generative grammar remains anchored to classical—though solid—models, whereas computational and experimental linguistics have endeavored to make the quantum leap.

## 4. A Cautious Conclusion

At the end of the roundtable on Hilbert's List for Syntax, I had the impression that everyone was satisfied with the current problems' formulation, even though there was little interaction between one problem and another. I believe this reflects the sentiment that pervaded among those observing generative grammar from the outside: those idealized problems do not fit with each other and appear to be complex quirks with negligible impact on the understanding of the language faculty. The emerging trend associated with the computational and experimental perspectives is that concrete linguistic facts must be investigated using experimentally solid methods and computationally robust tools. Exotic theoretical puzzles with funny names and acronyms—e.g., strong islands, ATB extraction, Complex NP constraint—that are unattested in naturalistic corpora can be safely ignored if our aim is to understand and perform realistic linguistic tasks.

In these pages, I have attempted to demonstrate that this represents a limited perspective for achieving descriptive and explanatory adequacy. However, I am convinced that the current form of the generative paradigm, namely Minimalism, does not effectively support this position. This is due to a lack of consistency at both the formal level, where crucial intuitions that restrict structure-building operations remain underspecified, leading to formalization issues, and at the empirical level, where most theoretical intuitions are supported only by a limited empirical domain. Once this domain is extended, it often conflicts with other intuitions—evaluation issues.

In the end, on the one side of the field, computational linguists depend on statistical predictions obtained from vast corpora and have shown that the core syntactic engine, PF, and LF, are effectively distinct only within the theoretical 'T-model'. To truly understand what a sentence means—crucial for tasks like machine translation or answering questions—it is essential to rely on robust machine learning methods, which are more solid than any formal theory on the market. On the other side of the field, experimental linguistics has refined its methods, significantly improving the observational capabilities and, ultimately, enhancing the analysis of nearly all sources of linguistic data, whether implicit or explicit, categorical or gradual. Both players adopted open science practices, including the sharing of data and methods, and increasingly relied on sophisticated statistical methodology. Advanced statistical methods, as anticipated, have contributed to the success of machine learning and inferential analysis more than anything else. Generative linguists are sitting on the bench, watching the game, laughing at some experimental

results—which seem occasionally to reinvent the wheel—, or expressing skepticism towards the inherent complexity of computational and statistical methods, as well as questioning their relevance as descriptive or explanatory theories. But they remained in the background. As Piantadosi provocatively said, this is "what happens when an academic field isolates itself from what should be complementary endeavours". While generative linguistics struggles to accommodate gradual judgments, online effects, and other kinds of implicit data, these are the daily bread of computational and psycholinguistic models.

As demonstrated in these pages, generative linguistics has had the opportunity—and the ability—to establish the level of complexity for linguistic puzzles that need solving. However, this turned into a Pyrrhic victory: in focusing on the 'speck' in vLLMs' eye, they failed to recognize the 'log' in their own, which, in my opinion, is the standard MG model's inability to address incrementality, together with a lack of specification of fundamental constraints, and the absence of a shared evaluation benchmark.

While vLLMs are arguably overrated as linguistic theories, the methodology proposed by Wilcox and colleagues (Wilcox, Futrell, and Levy 2023) represents an appropriate approach to testing them. By training different architectures from scratch on various plausible datasets (Warstadt et al. 2023), while avoiding fine-tuning, we may discover that the majority of the contrasts we aim to capture are indeed learnable. This outcome would offer evidence in favor of the PoS hypothesis, proving that the architectural intuitions explanatorily support language learnability. So far, the vLLMs tested are the only ones that perform properly on shared benchmarks such as SyntaxGym, CoLA, or BLiMP. Their dimension might be an issue, but only when a smaller model would obtain a comparable level of accuracy on these tests. In this respect, these vLLMs are, in fact, really the best theories on the market, i.e. observationally more adequate than any MG.

It is widely recognized that new ideas—whether radical or minimal—are often supported by limited data, which computational and psycho-/neuro-linguists might deem marginal. The argument typically proposes a novel theoretical component, Y, needed to capture a distinction between evidence A and B—a distinction previously unaccounted for, thereby improving observational adequacy at the cost of increased size. Alternatively, it suggests replacing components Y and X with Z to accommodate the same dataset, thus enhancing descriptive adequacy through a reduction in theory size. Despite the innovation introduced by these ideas, their successful integration depends on addressing formalization and evaluation issues. The updated framework must undergo testing not only against the limited data that indicated the need for a novel component but also across the entire dataset. Such comprehensive evaluation is crucial to ensure that the new intuition neither overgeneralizes nor undergeneralizes.

I think the original sin of most generative linguists is that they have gotten used to incomplete pseudo-formalizations and data fragment explanations.

Making meaningful comparisons poses a significant challenge without addressing both the consistency and completeness of the theory, related to formalization issues, and establishing a reference dataset, pertaining to evaluation concerns. This situation has led to the rise of "Personal Minimalisms", characterized by predominant subjective interpretations. Consequently, generative linguists may continue to superficially dismiss "murky judgments". Meanwhile, psycho- and computational linguists could potentially uncover significant insights hidden within these ambiguities. We are well aware that not everything that glitters is gold. Ultimately, the most significant contribution that a generative linguist can provide is a linguistic minimal contrast challenging a specific theoretical assumption or the performance of a vLLM. Successfully incorporating this new contrast into a shared dataset, which any (r)evolutionary explicit formalism must confront, would represent quite a considerable accomplishment in my opinion.

If this does not happen, I fear that it might be the end of generative linguistics as we know it (but I feel fine).

## *Abbreviations*

ATB: Across-The-Board extraction (Williams 1977)

BLiMP: Benchmark of Linguistic Minimal Pairs for English (Warstadt, Parrish, et al. 2020)

CI: Conceptual-Intentional interface

CoLA: Corpus of Linguistic Acceptability (Warstadt, Singh, and Bowman 2018)

CP: Complementized Phrase

DP: Determiner Phrase

GU20: Greenberg's Universal 20 (Cinque 2005)

LCA: Linear Correspondence Axiom (Kayne 1994)

LF: Logical Form

LSTM: Long-Short Term Memory networks (Sepp Hochreiter and Schmidhuber 1997)

MG: Minimalist Grammar

MDL: Minimum Description Length (Rissanen 1978)

MSGS: Mixed Signals Generalization Set (Warstadt, Zhang, et al. 2020)

NP: Noun Phrase

PF: Phonetic Form

QR: Quantifier Raising (May 1985)

SM: Sensory-Motor interface

SRN: Simple Recurrent Networks (Elman 1990)

SoA: Stage of Acquisition

vLLM: very Large Language Models

VP: Verb Phrase

## References

Abels, Klaus, and Ad Neeleman. 2012. "Linear Asymmetries and the LCA: Linear Asymmetries and the LCA." *Syntax* 15 (1): 25–74. https://doi.org/10.1111/j.1467-9612.2011.00163.x.

Baker, Mark C. 2001. *The Atoms of Language*. 1st ed. New York: Basic Books.

Baltin, Mark. 2017. "Extraposition." In *The Wiley Blackwell Companion to Syntax, Second Edition*, edited by Martin Everaert and Henk C. van Riemsdijk, 1–33. Hoboken, NJ, USA: John Wiley & Sons, Inc. https://doi.org/10.1002/9781118358733.wbsyncom111.

Barton, G. Edward, Robert C. Berwick, and Eric Sven Ristad. 1987. *Computational Complexity and Natural Language*. Computational Models of Cognition and Perception. Cambridge, Mass: MIT Press.

Bates, Elizabeth, Jeffrey Elman, Mark H. Johnson, Annette Karmiloff-Smith, Domenico Parisi, and Kim Plunkett. 1996. *Rethinking Innateness: A Connectionist Perspective on Development*. The MIT Press. https://doi.org/10.7551/mitpress/5929.001.0001.

Beghelli, Filippo, and Tim Stowell. 1997. "Distributivity and Negation: The Syntax of Each and Every." In *Ways of Scope Taking*, edited by Anna Szabolcsi, 65:71–107. Studies in Linguistics and Philosophy. Dordrecht: Springer Netherlands. https://doi.org/10.1007/978-94-011-5814-5_3.

Belletti, Adriana. 2004. *Structures and Beyond: The Cartography of Syntactic Structures, Volume 3*. Oxford University Press.

Bever, Thomas G. 1970. "The Cognitive Basis for Linguistic Structures." *Cognition and the Development of Language*.

Bever, Thomas G., and David J. Townsend. 2001. "Some Sentences on Our Consciousness of Sentences." In *Language, Brain, and Cognitive Development: Essays in Honor of Jacques Mehler*, edited by Emmanuel Dupoux, 143–55. MIT Press.

Bianchi, Valentina, and Cristiano Chesi. 2014. "Subject Islands, Reconstruction, and the Flow of the Computation." *Linguistic Inquiry*, 525–69. https://doi.org/10.1162/LING_a_00166.

Bloom, Paul A., and Ira Fischler. 1980. "Completion Norms for 329 Sentence Contexts." *Memory & Cognition* 8 (6): 631–42. https://doi.org/10.3758/BF03213783.

Bock, J. Kathryn. 1986. "Meaning, Sound, and Syntax: Lexical Priming in Sentence Production." *Journal of Experimental Psychology: Learning, Memory, and Cognition* 12 (4): 575–86. https://doi.org/10.1037/0278-7393.12.4.575.

Bošković, Željko. 2016. "Introduction." *The Linguistic Review* 33 (1): 1–16. https://doi.org/10.1515/tlr-2015-0012.

Brennan, Jonathan R., Edward P. Stabler, Sarah E. Van Wagenen, Wen-Ming Luh, and John T. Hale. 2016. "Abstract Linguistic Structure Correlates with Temporal Activity during Naturalistic Comprehension." *Brain and Language* 157–158 (June):81–94. https://doi.org/10.1016/j.bandl.2016.04.008.

Brown, Tom B., Benjamin Mann, Nick Ryder, Melanie Subbiah, Jared Kaplan, Prafulla Dhariwal, Arvind Neelakantan, et al. 2020. "Language Models Are Few-Shot Learners." *arXiv:2005.14165 [Cs]*, July. http://arxiv.org/abs/2005.14165.

Brunato, Dominique, Cristiano Chesi, Felice Dell'Orletta, Simonetta Montemagni, Giulia Venturi, and Roberto Zamparelli. 2020. "AcCompl-It@ EVALITA2020: Overview of the Acceptability & Complexity Evaluation Task for Italian." In *Proceedings of Seventh Evaluation Campaign of Natural Language Processing and Speech Tools for Italian. Final Workshop (EVALITA 2020), Online. CEUR. Org*.

Chaitin, Gregory J. 1969. "On the Simplicity and Speed of Programs for Computing Infinite Sets of Natural Numbers." *Journal of the ACM* 16 (3): 407–22. https://doi.org/10.1145/321526.321530.

Charpentier, Lucas Georges Gabriel, and David Samuel. 2023. "Not All Layers Are Equally as Important: Every Layer Counts BERT." In *Proceedings of the BabyLM Challenge at the 27th Conference on Computational Natural Language Learning*, 210–24. Singapore: Association for Computational Linguistics. https://doi.org/10.18653/v1/2023.conll-babylm.20.

Chesi, Cristiano. 2021. "Expectation-Based Minimalist Grammars." *arXiv:2109.13871 [Cs]*, September. http://arxiv.org/abs/2109.13871.

———. 2023. "Parameters of Cross-Linguistic Variation in Expectation-Based Minimalist Grammars (e-MGs)." *Italian Journal of Computational Linguistics* 9 (1): 21.

Chesi, Cristiano, and Andrea Moro. 2015. "The Subtle Dependency between Competence and Performance." *MIT WORKING PAPERS IN LINGUISTICS* 77:33–46.

Chesi, Cristiano, Francesco Vespignani, and Roberto Zamparelli. to appear. "Large Language Models under Evaluation: An Acceptability, Complexity and Coherence Assessment in Italian." *Italian Journal of Computational Linguistics*.

Chomsky, N. 1956. "Three Models for the Description of Language." *IEEE Transactions on Information Theory* 2 (3): 113–24. https://doi.org/10.1109/TIT.1956.1056813.

Chomsky, Noam. 1959. "A Review of B. F. Skinner's Verbal Behavior." *Language* 35 (1): 26. https://doi.org/10.2307/411334.

———. 1964. *Current Issues in Linguistic Theory*. Berlin: De Gruyter.

———. 1965. *Aspects of the Theory of Syntax*. Vol. 11. Cambridge, MA: MIT Press.

———. 1981. *Lectures on Government and Binding: The Pisa Lectures*. 9. Walter de Gruyter.

———. 1986. *Knowledge of Language: Its Nature, Origin, and Use*. Convergence. New York: Praeger.

———. 1995. *The Minimalist Program*. Cambridge, MA: MIT Press.

———. 2001. "Derivation by Phase." In *Ken Hale: A Life in Language*, edited by Michael Kenstowicz, 1–52. 36. Cambridge (MA): MIT Press.

———. 2008. "On Phases." In *Foundational Issues in Linguistic Theory: Essays in Honor of Jean-Roger Vergnaud*, edited by Robert Freidin, Carlos P. Otero, and Maria Luisa Zubizarreta, 45:133–66. Cambridge, MA: MIT Press.

———. 2013. "Problems of Projection." *Lingua* 130:33–49.

———. 2015. "Problems of Projection: Extensions." In *Linguistik Aktuell/Linguistics Today*, edited by Elisa Di Domenico, Cornelia Hamann, and Simona Matteini, 223:1–16. Amsterdam: John Benjamins Publishing Company. https://doi.org/10.1075/la.223.01cho.

———. 2021. "Simplicity and the Form of Grammars." *Journal of Language Modelling* 9 (1). https://doi.org/10.15398/jlm.v9i1.257.

Chomsky, Noam, and Howard Lasnik. 1977. "Filters and Control." *Linguistic Inquiry* 8 (3): 425–504.

Chomsky, Noam, Ian Roberts, and Jeffrey Watumull. 2023. "Noam Chomsky: The False Promise of ChatGPT." *New York Times*, 2023.

Chomsky, Noam, T. Daniel Seely, Robert C. Berwick, Sandiway Fong, M. A. C. Huybregts, Hisatsugu Kitahara, Andrew McInnerney, and Yushi Sugimoto. 2023. *Merge and the Strong Minimalist Thesis*. 1st ed. Cambridge University Press. https://doi.org/10.1017/9781009343244.

Chowdhury, Shammur Absar, and Roberto Zamparelli. 2018. "RNN Simulations of Grammaticality Judgments on Long-Distance Dependencies." In *Proceedings of the 27th International Conference on Computational Linguistics*, 133–44. Santa Fe, New Mexico, USA: Association for Computational Linguistics. https://aclanthology.org/C18-1012.

Cinque, Guglielmo. 1999. *Adverbs and Functional Heads: A Cross-Linguistic Perspective*. Oxford (UK): Oxford University Press.

———. 2002. *Functional Structure in DP and IP: The Cartography of Syntactic Structures, Volume 1*. Oxford University Press.

———. 2005. "Deriving Greenberg's Universal 20 and Its Exceptions." *Linguistic Inquiry* 36 (3): 315–32. https://doi.org/10.1162/0024389054396917.

Clark, Alexander, and Shalom Lappin. 2010. "Computational Learning Theory and Language Acquisition." *Philosophy of Linguistics*, 445–75.

Collins, Chris, Richard Kayne, and Hilda Koopman. 2009. "Syntactic Structures of the World's Languages (SSWL)." 2009. https://terraling.com/groups/7.

Collins, Chris, and Edward Stabler. 2016. "A Formalization of Minimalist Syntax." *Syntax* 19 (1): 43–78. https://doi.org/10.1111/synt.12117.

Crain, Stephen, and Mineharu Nakayama. 1987. "Structure Dependence in Grammar Formation." *Language* 63 (3): 522. https://doi.org/10.2307/415004.

De Santo, Aniello. 2019. "Testing a Minimalist Grammar Parser On." In *Proceedings of the Workshop on Cognitive Modeling and Computational Linguistics*, 93–104. Minneapolis, Minnesota: Association for Computational Linguistics. https://doi.org/10.18653/v1/W19-2911.

Dentella, Vittoria, Fritz Günther, and Evelina Leivada. 2023. "Systematic Testing of Three Language Models Reveals Low Language Accuracy, Absence of Response Stability, and a Yes-Response Bias." *Proceedings of the National Academy of Sciences* 120 (51): e2309583120. https://doi.org/10.1073/pnas.2309583120.

Dryer, Matthew, and Martin Haspelmath. 2022. "The World Atlas of Language Structures Online." Zenodo. https://doi.org/10.5281/ZENODO.7385533.

Elman, Jeffrey L. 1990. "Finding Structure in Time." *Cognitive Science* 14 (2): 179–211. https://doi.org/10.1207/s15516709cog1402_1.

———. 1993. "Learning and Development in Neural Networks: The Importance of Starting Small." *Cognition* 48 (1): 71–99. https://doi.org/10.1016/0010-0277(93)90058-4.

Epstein, Samuel David, Erich M. Groat, Ruriko Kawashima, and Hisatsugu Kitahara, eds. 1998. *A Derivational Approach to Syntactic Relations*. New York: Oxford University Press.

Evans, Lyndon. 2007. "The Large Hadron Collider." *New Journal of Physics* 9 (9): 335–335. https://doi.org/10.1088/1367-2630/9/9/335.

Fodor, Jerry A. 1983. *The Modularity of Mind: An Essay on Faculty Psychology*. Cambridge, MA: MIT Press.

Forster, Kenneth I., Christine Guerrera, and Lisa Elliot. 2009. "The Maze Task: Measuring Forced Incremental Sentence Processing Time." *Behavior Research Methods* 41 (1): 163–71. https://doi.org/10.3758/BRM.41.1.163.

Fox, Danny, and Jon Nissenbaum. 1999. "Extraposition and Scope: A Case for Overt QR." In *Proceedings of the 18th West Coast Conference on Formal Linguistics*, 18:132–44.

Frampton, John, and Sam Gutmann. 2002. "Crash-Proof Syntax." In *Derivation and Explanation in the Minimalist Program*, edited by Samuel David Epstein and T. Daniel Seely, 1st ed., 90–105. Wiley. https://doi.org/10.1002/9780470755662.ch5.

Frank, Robert. 1990. "Licensing and Tree Adjoining Grammar in Government Binding Parsing." In *28th Annual Meeting of the Association for Computational Linguistics*, 111–18.

———. 2002. *Phrase Structure Composition and Syntactic Dependencies*. Current Studies in Linguistics Series 38. Cambridge, MA: MIT Press.

Friedmann, Naama, Adriana Belletti, and Luigi Rizzi. 2009. "Relativized Relatives: Types of Intervention in the Acquisition of A-Bar Dependencies." *Lingua* 119 (1): 67–88.

Futrell, Richard, Edward Gibson, and Roger P. Levy. 2020. "Lossy-Context Surprisal: An Information-Theoretic Model of Memory Effects in Sentence Processing." *Cognitive Science* 44 (3). https://doi.org/10.1111/cogs.12814.

Futrell, Richard, and Roger Levy. 2017. "Noisy-Context Surprisal as a Human Sentence Processing Cost Model." In *Proceedings of the 15th Conference of the European Chapter of the Association for Computational Linguistics: Volume 1, Long Papers*, 688–98.

Gianollo, Chiara, Cristina Guardiano, and Giuseppe Longobardi. 2008. "Three Fundamental Issues in Parametric Linguistics." In *Linguistik Aktuell/Linguistics Today*, edited by Theresa Biberauer, 132:109–42. Amsterdam: John Benjamins Publishing Company. https://doi.org/10.1075/la.132.05gia.

Ginsburg, Jason, and Sandiway Fong. 2019. "Combining Linguistic Theories in a Minimalist Machine." In *Minimalist Parsing*, by Jason Ginsburg and Sandiway Fong, 39–68. Oxford University Press. https://doi.org/10.1093/oso/9780198795087.003.0003.

Gold, E Mark. 1967. "Language Identification in the Limit." *Information and Control* 10 (5): 447–74. https://doi.org/10.1016/S0019-9958(67)91165-5.

Graf, Thomas, James Monette, and Chong Zhang. 2017. "Relative Clauses as a Benchmark for Minimalist Parsing." *Journal of Language Modelling* 5 (1). https://doi.org/10.15398/jlm.v5i1.157.

Grillo, Nino. 2008. *Generalized Minimality: Syntactic Underspecification in Broca's Aphasia*. LOT.

Grünwald, Peter D. 2007. *The Minimum Description Length Principle*. Adaptive Computation and Machine Learning. Cambridge, MA: MIT Press.

Guardiano, Cristina, and Giuseppe Longobardi. 2016. "Parameter Theory and Parametric Comparison." In *The Oxford Handbook of Universal Grammar*, by Cristina Guardiano and Giuseppe Longobardi, edited by Ian Roberts, 376–98. Oxford University Press. https://doi.org/10.1093/oxfordhb/9780199573776.013.16.

Guardiano, Cristina, Giuseppe Longobardi, Guido Cordoni, and Paola Crisma. 2020. "Formal Syntax as a Phylogenetic Method." In *The Handbook of Historical Linguistics*, edited by Richard D. Janda, Brian D. Joseph, and Barbara S. Vance, 1st ed., 145–82. Wiley. https://doi.org/10.1002/9781118732168.ch7.

Guasti, Maria Teresa. 2017. *Language Acquisition: The Growth of Grammar*. Cambridge, MA: MIT press.

Hale, John. 2001. "A Probabilistic Earley Parser as a Psycholinguistic Model." In *Second Meeting of the North American Chapter of the Association for Computational Linguistics*. https://aclanthology.org/N01-1021.

———. 2011. "What a Rational Parser Would Do." *Cognitive Science* 35 (3): 399–443.

———. 2016. "Information-theoretical Complexity Metrics." *Language and Linguistics Compass* 10 (9): 397–412. https://doi.org/10.1111/lnc3.12196.

Hart, Betty, and Todd R. Risley. 1992. "American Parenting of Language-Learning Children: Persisting Differences in Family-Child Interactions Observed in Natural Home Environments." *Developmental Psychology* 28 (6): 1096–1105. https://doi.org/10.1037/0012-1649.28.6.1096.

Hochreiter, S., Y. Bengio, P. Frasconi, and J. Schmidhuber. 2001. "Gradient Flow in Recurrent Nets: The Difficulty of Learning Long-Term Dependencies." In *A Field Guide to Dynamical Recurrent Neural Networks*, edited by S. C. Kremer and J. F. Kolen. IEEE Press.

Hochreiter, Sepp, and Jürgen Schmidhuber. 1997. "Long Short-Term Memory." *Neural Computation* 9 (8): 1735–80.

Hsu, Anne S., and Nick Chater. 2010. "The Logical Problem of Language Acquisition: A Probabilistic Perspective." *Cognitive Science* 34 (6): 972–1016. https://doi.org/10.1111/j.1551-6709.2010.01117.x.

Hsu, Anne S., Nick Chater, and Paul Vitányi. 2013. "Language Learning From Positive Evidence, Reconsidered: A Simplicity-Based Approach." *Topics in Cognitive Science* 5 (1): 35–55. https://doi.org/10.1111/tops.12005.

Hu, Jennifer, Jon Gauthier, Peng Qian, Ethan Wilcox, and Roger Levy. 2020. "A Systematic Assessment of Syntactic Generalization in Neural Language Models." In *Proceedings of the 58th Annual Meeting of the Association for Computational Linguistics*, edited by Dan Jurafsky, Joyce Chai, Natalie Schluter, and Joel Tetreault, 1725–44. Online: Association for Computational Linguistics. https://doi.org/10.18653/v1/2020.acl-main.158.

Huang, C.-T. James. 1982. "Logical Relations in Chinese and the Theory of Grammar." Cambridge (MA): MIT.

Kaplan, Jared, Sam McCandlish, Tom Henighan, Tom B. Brown, Benjamin Chess, Rewon Child, Scott Gray, Alec Radford, Jeffrey Wu, and Dario Amodei. 2020. "Scaling Laws for Neural Language Models." https://doi.org/10.48550/ARXIV.2001.08361.

Katzir, Roni. 2023. "Why Large Language Models Are Poor Theories of Human Linguistic Cognition. A Reply to Piantadosi (2023)." Lingbuzz. lingbuzz/007190.

Kayne, Richard S. 1994. *The Antisymmetry of Syntax*. Linguistic Inquiry Monographs 25. Cambridge, MA: MIT Press.

Kobele, Gregory M. 2023. "Minimalist Grammars and Decomposition." In *The Cambridge Handbook of Minimalism*, edited by Grohmann Kleanthes K. and Evelina Leivada. Cambridge (UK): Cambridge University Press.

Kolmogorov, Andrey N. 1963. "On Tables of Random Numbers." *Sankhyā: The Indian Journal of Statistics, Series A (1961-2002)* 25 (4): 369–76.

Levesque, Hector J. 2014. "On Our Best Behaviour." *Artificial Intelligence* 212 (July):27–35. https://doi.org/10.1016/j.artint.2014.03.007.

Levy, Roger. 2008. "Expectation-Based Syntactic Comprehension." *Cognition* 106 (3): 1126–77.

Li, Jixing, and John Hale. 2019. "Grammatical Predictors for fMRI Time-Courses." In *Minimalist Parsing*, by Robert C. Berwick and Edward P. Stabler, 159–73. Oxford University Press. https://doi.org/10.1093/oso/9780198795087.003.0007.

Li, Ming, and Paul Vitányi. 2008. *An Introduction to Kolmogorov Complexity and Its Applications*. Texts in Computer Science. New York, NY: Springer New York. https://doi.org/10.1007/978-0-387-49820-1.

Linzen, Tal, Emmanuel Dupoux, and Yoav Goldberg. 2016. "Assessing the Ability of LSTMs to Learn Syntax-Sensitive Dependencies." *Transactions of the Association for Computational Linguistics* 4 (December):521–35. https://doi.org/10.1162/tacl_a_00115.

Marr, David. 1982. *Vision: A Computational Investigation into the Human Representation and Processing of Visual Information*. W.H. Freeman. San Francisco (CA): Freeman.

Marr, David, and Tomaso Poggio. 1976. "From Understanding Computation to Understanding Neural Circuitry." USA: Massachusetts Institute of Technology.

Marvin, Rebecca, and Tal Linzen. 2018. "Targeted Syntactic Evaluation of Language Models." In *Proceedings of the 2018 Conference on Empirical Methods in Natural Language Processing*, 1192–1202. Brussels, Belgium: Association for Computational Linguistics. https://doi.org/10.18653/v1/D18-1151.

May, Robert. 1985. *Logical Form: Its Structure and Derivation*. Vol. 12. Cambridge, MA: MIT Press.

Michaelis, Jens. 2001. "Derivational Minimalism Is Mildly Context–Sensitive." In *Logical Aspects of Computational Linguistics*, edited by Michael Moortgat, 2014:179–98. Lecture Notes in Computer Science. Berlin, Heidelberg: Springer Berlin Heidelberg. https://doi.org/10.1007/3-540-45738-0_11.

Miller, George A., and Noam Chomsky. 1963. "Finitary Models of Language Users." In *Handbook of Mathematical Psychology*, edited by D. Luce, 2–419. John Wiley & Sons.

Momma, Shota, and Colin Phillips. 2018. "The Relationship Between Parsing and Generation." *Annual Review of Linguistics* 4 (1): 233–54. https://doi.org/10.1146/annurev-linguistics-011817-045719.

Moro, Andrea. 2023. "Embodied syntax: Impossible languages and the irreducible difference between humans and machines." *Sistemi intelligenti*, no. 2, 321–28. https://doi.org/10.1422/108132.

Moro, Andrea, Matteo Greco, and Stefano F. Cappa. 2023. "Large Languages, Impossible Languages and Human Brains." *Cortex* 167 (October):82–85. https://doi.org/10.1016/j.cortex.2023.07.003.

Nivre, Joakim, Željko Agić, Lars Ahrenberg, Lene Antonsen, Maria Jesus Aranzabe, Masayuki Asahara, Luma Ateyah, et al. 2017. "Universal Dependencies 2.1."

Nosengo, Nicola. 2014. *I robot ci guardano: aerei senza pilota, chirurghi a distanza e automi solidali*. Bologna: Zanichelli.

Oerter, Robert. 2006. *The Theory of Almost Everything: The Standard Model, the Unsung Triumph of Modern Physics*. New York: Pi Press.

OpenAI. 2023. "GPT-4 Technical Report." arXiv. http://arxiv.org/abs/2303.08774.

Phillips, Colin. 1996. "Order and Structure." PhD Thesis, Massachusetts Institute of Technology.

Piantadosi, ST. 2023. "Modern Language Models Refute Chomsky's Approach to Language." *Lingbuzz Preprint, Lingbuzz* 7180.

Quinlan, Philip T., ed. 2004. *Connectionist Models of Development*. 0 ed. Psychology Press. https://doi.org/10.4324/9780203494028.

Radford, Alec, Karthik Narasimhan, Tim Salimans, Ilya Sutskever, and others. 2018. "Improving Language Understanding by Generative Pre-Training."

Reinhart, Tanya. 1976. "The Syntactic Domain of Anaphora." Cambridge (MA): Massachusetts Institute of Technology.

Riesenhuber, Maximilian, and Tomaso Poggio. 1999. "Hierarchical Models of Object Recognition in Cortex." *Nature Neuroscience* 2 (11): 1019–25. https://doi.org/10.1038/14819.

Rissanen, Jorma. 1978. "Modeling by Shortest Data Description." *Automatica* 14 (5): 465–71. https://doi.org/10.1016/0005-1098(78)90005-5.

———. 1987. "Stochastic Complexity." *Journal of the Royal Statistical Society: Series B (Methodological)* 49 (3): 223–39. https://doi.org/10.1111/j.2517-6161.1987.tb01694.x.

Rizzi, Luigi. 1990. *Relativized Minimality*. Linguistic Inquiry Monographs 16. Cambridge, MA: MIT Press.

———. 1997. "The Fine Structure of the Left Periphery." In *Elements of Grammar*, edited by Liliane Haegeman, 281–337. Kluwer International Handbooks of Linguistics. Dordrecht: Springer Netherlands. https://doi.org/10.1007/978-94-011-5420-8_7.

———, ed. 2004. *The Structure of CP and IP*. Oxford Studies in Comparative Syntax, v. 2. Oxford ; New York: Oxford University Press.

———. 2016. "Labeling, Maximality and the Head–Phrase Distinction." *The Linguistic Review* 33 (1): 103–27.

Rizzi, Luigi, and Guglielmo Cinque. 2016. "Functional Categories and Syntactic Theory." *Annual Review of Linguistics* 2 (1): 139–63. https://doi.org/10.1146/annurev-linguistics-011415-040827.

Roberts, Ian. 2017. "The Final-over-Final Condition in DP: Universal 20 and the Nature of Demonstratives." In *The Final-over-Final Condition: A Syntactic Universal*, edited by Michelle

Sheehan, Theresa Biberauer, Ian Roberts, and Anders Holmberg, 76:151. Cambridge (MA): MIT Press.

———. 2019. *Parameter Hierarchies and Universal Grammar*. 1st ed. Oxford University Press. https://doi.org/10.1093/oso/9780198804635.001.0001.

Ross, John Robert. 1967. "Constraints on Variables in Syntax." Cambridge (MA): MIT.

Rumelhart, David E., James L. McClelland, and PDP Research Group, eds. 1999. *Parallel Distributed Processing. 1: Foundations*. 12. print. Cambridge, Mass: MIT Pr.

Siegelman, Noam, Sascha Schroeder, Cengiz Acartürk, Hee-Don Ahn, Svetlana Alexeeva, Simona Amenta, Raymond Bertram, et al. 2022. "Expanding Horizons of Cross-Linguistic Research on Reading: The Multilingual Eye-Movement Corpus (MECO)." *Behavior Research Methods* 54 (6): 2843–63. https://doi.org/10.3758/s13428-021-01772-6.

Solomonoff, Ray. 1960. *A Preliminary Report on a General Theory of Inductive Inference*. AFOSR TN-60-1459. United States Air Force, Office of Scientific Research. https://books.google.it/books?id=SuTHtgAACAAJ.

Sprouse, Jon, and Diogo Almeida. 2017. "Design Sensitivity and Statistical Power in Acceptability Judgment Experiments." *Glossa* 2 (1): 1–32. https://doi.org/DOI: https://doi.org/10.5334/gjgl.236.

Sprouse, Jon, and Norbert Hornstein, eds. 2013. *Experimental Syntax and Island Effects*. 1st ed. Cambridge University Press. https://doi.org/10.1017/CBO9781139035309.

Stabler, Edward. 1997. "Derivational Minimalism." In *Logical Aspects of Computational Linguistics*, edited by Christian Retoré, 68–95. Berlin, Heidelberg: Springer Berlin Heidelberg.

———. 2011. "Computational Perspectives on Minimalism." In *The Oxford Handbook of Linguistic Minimalism*, edited by Cedric Boeckx. Oxford University Press. https://doi.org/10.1093/oxfordhb/9780199549368.013.0027.

———. 2013. "Two Models of Minimalist, Incremental Syntactic Analysis." *Topics in Cognitive Science* 5 (3): 611–33. https://doi.org/10.1111/tops.12031.

Starke, Michal. 2001. "Move Dissolves into Merge: A Theory of Locality." PhD Thesis, Université de Genève.

Steuer, Julius, Marius Mosbach, and Dietrich Klakow. 2023. "Large GPT-like Models Are Bad Babies: A Closer Look at the Relationship between Linguistic Competence and Psycholinguistic Measures." In *Proceedings of the BabyLM Challenge at the 27th Conference on Computational Natural Language Learning*, 114–29. Singapore: Association for Computational Linguistics. https://doi.org/10.18653/v1/2023.conll-babylm.12.

Taylor, Wilson L. 1953. "'Cloze Procedure': A New Tool for Measuring Readability." *Journalism Quarterly* 30 (4): 415–33. https://doi.org/10.1177/107769905303000401.

Trotta, Daniela, Raffaele Guarasci, Elisa Leonardelli, and Sara Tonelli. 2021. "Monolingual and Cross-Lingual Acceptability Judgments with the Italian CoLA Corpus." In *Findings of the Association for Computational Linguistics: EMNLP 2021*, 2929–40. Punta Cana, Dominican Republic: Association for Computational Linguistics. https://doi.org/10.18653/v1/2021.findings-emnlp.250.

Vaswani, Ashish, Noam Shazeer, Niki Parmar, Jakob Uszkoreit, Llion Jones, Aidan N. Gomez, Lukasz Kaiser, and Illia Polosukhin. 2017. "Attention Is All You Need." *arXiv:1706.03762 [Cs]*, December. http://arxiv.org/abs/1706.03762.

Vermeerbergen, Myriam, Lorraine Leeson, and Onno Alex Crasborn, eds. 2007. *Simultaneity in Signed Languages: Form and Function*. Amsterdam Studies in the Theory and History of Linguistic Science, v. 281. Amsterdam ; Philadelphia: John Benjamins.

Warstadt, Alex, Aaron Mueller, Leshem Choshen, Ethan Wilcox, Chengxu Zhuang, Juan Ciro, Rafael Mosquera, et al. 2023. "Findings of the BabyLM Challenge: Sample-Efficient Pretraining on Developmentally Plausible Corpora." In *Proceedings of the BabyLM Challenge at the 27th Conference on Computational Natural Language Learning*, 1–6. Singapore: Association for Computational Linguistics. https://doi.org/10.18653/v1/2023.conll-babylm.1.

Warstadt, Alex, Alicia Parrish, Haokun Liu, Anhad Mohananey, Wei Peng, Sheng-Fu Wang, and Samuel R. Bowman. 2020. "BLiMP: The Benchmark of Linguistic Minimal Pairs for English." *Transactions of the Association for Computational Linguistics* 8 (December):377–92. https://doi.org/10.1162/tacl_a_00321.

Warstadt, Alex, Amanpreet Singh, and Samuel R Bowman. 2018. "Neural Network Acceptability Judgments." *arXiv Preprint arXiv:1805.12471*.

Warstadt, Alex, Yian Zhang, Xiaocheng Li, Haokun Liu, and Samuel R. Bowman. 2020. "Learning Which Features Matter: RoBERTa Acquires a Preference for Linguistic Generalizations (Eventually)." In *Proceedings of the 2020 Conference on Empirical Methods in Natural Language Processing (EMNLP)*, 217–35. Online: Association for Computational Linguistics. https://doi.org/10.18653/v1/2020.emnlp-main.16.

Wilcox, Ethan, Richard Futrell, and Roger Levy. 2023. "Using Computational Models to Test Syntactic Learnability." *Linguistic Inquiry*, April, 1–44. https://doi.org/10.1162/ling_a_00491.

Wilcox, Ethan, Roger Levy, Takashi Morita, and Richard Futrell. 2018. "What Do RNN Language Models Learn about Filler-Gap Dependencies?" arXiv. http://arxiv.org/abs/1809.00042.

Williams, Edwin S. 1977. "Discourse and Logical Form." *Linguistic Inquiry* 8 (1): 101–39.

Yang, Charles D. 2016. *The Price of Linguistic Productivity: How Children Learn to Break the Rules of Language*. Cambridge, MA: MIT Press.

Yang, Yuan, and Steven T. Piantadosi. 2022. "One Model for the Learning of Language." *Proceedings of the National Academy of Sciences* 119 (5): e2021865119. https://doi.org/10.1073/pnas.2021865119.

Zhang, Yian, Alex Warstadt, Haau-Sing Li, and Samuel R. Bowman. 2020. "When Do You Need Billions of Words of Pretraining Data?" arXiv. http://arxiv.org/abs/2011.04946.